\documentclass{article}

\PassOptionsToPackage{numbers}{natbib}

\usepackage[main, final]{neurips_2026}

\usepackage[utf8]{inputenc}
\usepackage[T1]{fontenc}
\usepackage{hyperref}
\usepackage{url}
\usepackage{booktabs}
\usepackage{amsfonts}
\usepackage{nicefrac}
\usepackage{microtype}
\usepackage{xcolor}

\usepackage{multirow}
\usepackage{graphicx}
\usepackage{array}
\usepackage{amsmath}
\usepackage{amssymb}
\usepackage{caption}
\usepackage{subcaption}
\usepackage{float}
\usepackage{placeins}

\newcommand{\eg}{\emph{e.g.}}

\definecolor{mypink_dark}{RGB}{199, 21, 133}

\newcommand{\TODO}[1]{\textbf{\color{red}[TODO: #1]}}

\definecolor{addcolor}{RGB}{0, 90, 200}

\renewcommand{\TODO}[1]{}

\newif\ifrevmark
\revmarkfalse
\newcommand{\rev}[1]{\ifrevmark{\color{red}#1}\else#1\fi}
\newcommand{\revcolor}{\ifrevmark\color{red}\fi}

\hypersetup{hidelinks}

\title{FlowHOI: Flow-based Semantics-Grounded Generation of Hand-Object Interactions for Dexterous Robot Manipulation}

\author{%
  Huajian Zeng$^{1}$
  \quad
  Lingyun Chen$^{2}$
  \quad
  Jiaqi Yang$^{1,3}$
  \quad
  Yuantai Zhang$^{1}$ \\[1ex]
  \bfseries
  Fan Shi$^{4}$
  \quad
  Peidong Liu$^{5}$
  \quad
  Xingxing Zuo$^{1,*}$
  \\[2ex]
  \normalfont
  $^{1~}$Mohamed bin Zayed University of Artificial Intelligence \\
  $^{2~}$Technical University of Munich
  \quad
  $^{3~}$ShanghaiTech University \\
  $^{4~}$National University of Singapore
  \quad
  $^{5~}$Westlake University
}

\begin{document}

\maketitle

\renewcommand{\thefootnote}{\fnsymbol{footnote}}
\footnotetext{Project page: \url{https://huajian-zeng.github.io/projects/flowhoi}}
\footnotetext[1]{Corresponding author. (Email: {\tt\small xingxing.zuo@mbzuai.ac.ae}) }
\renewcommand{\thefootnote}{\arabic{footnote}}
\setcounter{footnote}{0}

\begin{figure}[h]
  \centering
  \includegraphics[width=\linewidth]{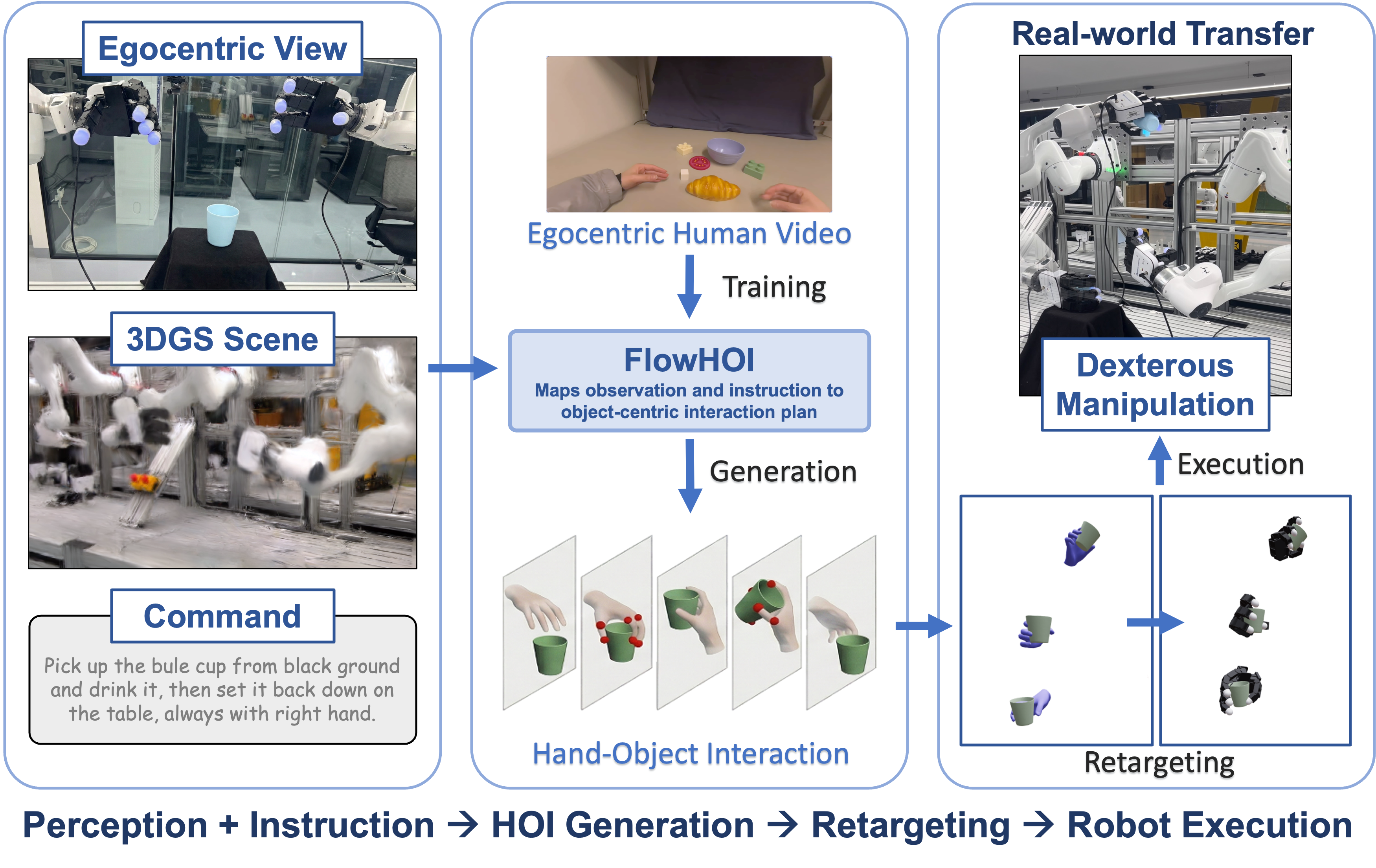}
  \caption{We present a method for generating hand-object interaction (HOI) motions conditioned on egocentric observation, text command, and 3D scene context.
  We first learn a grasping prior with HOI data extracted from large-scale egocentric videos, and then generate semantically grounded manipulation motions that respect language instructions as well as the surrounding 3D scene context and geometric constraints.
  The generated motions can be retargeted to robot hands for real-world execution.}
  \label{fig:teaser}
\end{figure}

\begin{abstract}
Recent vision-language-action (VLA) models can generate plausible end-effector motions, yet they often fail in long-horizon, contact-rich tasks because the underlying hand-object interaction (HOI) structure is not explicitly represented.
An embodiment-agnostic interaction representation that captures this structure would make manipulation behaviors easier to validate and transfer across robots.
We propose FlowHOI, a two-stage flow-matching framework that generates semantically grounded, temporally coherent HOI sequences, comprising hand poses, object poses, and hand-object contact states, conditioned on an egocentric observation, a language instruction, and a 3D Gaussian splatting (3DGS) scene reconstruction.
We decouple geometry-centric grasping from semantics-centric manipulation, conditioning the latter on compact 3D scene tokens and employing a motion-text alignment loss to semantically ground the generated interactions in both the physical scene layout and the language instruction.
To address the scarcity of high-fidelity HOI supervision, we introduce a reconstruction pipeline that recovers aligned hand-object trajectories and meshes from large-scale egocentric videos, yielding an HOI prior for robust generation.
Across the GRAB and HOT3D benchmarks, FlowHOI achieves the highest action recognition accuracy and a 1.7$\times$ higher physics-simulation success rate on GRAB than the strongest diffusion-based baseline, while delivering a 40$\times$ inference speedup.
%
We further demonstrate real-robot execution across four contact-rich, dexterous manipulation tasks, highlighting the effectiveness of retargeting our generated HOI representations to robotic dexterous hands, achieving significantly higher success rates than prior methods due to improved physical plausibility and semantic alignment.

\end{abstract}


\section{Introduction}
\label{sec:introduction}


Robotic manipulation in everyday household environments is fundamentally interaction-centric~\cite{Gibson1979-GIBTEA, mason2001mechanics, bicchi2000robotic}: tasks such as opening a lid, pouring from a container, or placing an object are not characterized by the robot's end-effector trajectory alone, but by the underlying context of interaction, namely where contact occurs given the scene geometry, how stable contact is preserved, and how the object's state evolves in a way that is semantically consistent with the language instruction and the scene affordances.

Recent vision-language-action (VLA) models~\cite{kim2024openvla,black2026pi0visionlanguageactionflowmodel,wang2025unified}, fine-tuned from large VLMs~\cite{alayrac2022flamingo,peng2023kosmos,beyer2024paligemma,li2025eagle}, generate plausible end-effector trajectories but struggle with contact-rich, long-horizon dexterous tasks~\cite{zhong2025dexgraspvla}.
This motivates a semantically grounded, embodiment-agnostic intermediate representation~\cite{pan2025omnimanip, xu2025a0} that explicitly encodes how contact is established and transitioned to induce language-specified state changes~\cite{jiang2021hand}, decoupling manipulation from robot-specific control and facilitating transfer across embodiments.

In this work, we propose FlowHOI, a two-stage flow-matching framework for producing semantically grounded hand-object interaction (HOI) sequences.
Given an initial egocentric observation, a language instruction, and a 3D Gaussian splatting (3DGS) scene reconstruction, FlowHOI generates temporally coherent and physically plausible HOI sequences comprising hand poses, object poses, and hand-object contact states, all anchored in the observed scene context and aligned with the language instruction.
The resulting HOI representation provides a natural interface for enforcing geometric and physical constraints and can be retargeted to downstream robotic dexterous manipulation\rev{, where it facilitates physically plausible execution}~\cite{hsu2025spot,lum2025crossing}.

Building such a model raises three central challenges:
%
%
\textbf{(i)~Geometric consistency \& semantic grounding.}
Long-horizon interactions are simultaneously constrained by 3D scene geometry and by the language instruction's semantics. \rev{Entangling both objectives in a single generator weakens contact establishment (Supp.~Sec.~\ref{sec:supp_single_stage}).} Inspired by how humans first establish a stable grasp before manipulating an object~\cite{johansson2009coding}, we decompose generation into a geometry-centric \emph{Grasping} stage that uses a pretrained grasping prior, followed by a semantics-centric \emph{Manipulation} stage conditioned on compact 3D scene tokens with a motion-text alignment loss.
\textbf{(ii)~Inference efficiency.}
Diffusion-based HOI generators~\cite{christen2024diffh2o,li2025latenthoi,ho2020denoising, song2020denoising} require tens to hundreds of denoising steps, taking 3--7\,s per sequence, which is prohibitive for real-time downstream planning.
We adopt conditional flow matching, reducing inference to 0.16\,s per sequence (up to 40$\times$ speedup) while maintaining temporally coherent and physically plausible generation.
\textbf{(iii)~Data scarcity.}
High-fidelity HOI supervision is scarce because hand-object interactions are high-dimensional, frequently occluded, and tightly coupled with contact dynamics~\cite{fu2026egograsp, xu2023egopca}.
We address this by introducing a reconstruction pipeline that recovers aligned hand-object trajectories and meshes from large-scale egocentric videos~\cite{hoque2025egodex}. \rev{We use the resulting data to pretrain the grasping stage, which improves grasp accuracy and the smoothness of the grasp-to-manipulation transition.}
%


To the best of our knowledge, FlowHOI is the \textit{first} method to formulate HOI generation as a unified, conditional flow-matching process in a semantics-grounded way. In summary, our contributions are:
\begin{itemize}
    \item We introduce a two-stage HOI generation framework that decouples geometry-centric grasping from semantics-centric manipulation and employs flow matching for efficient generation, achieving up to 40$\times$ speedup over prior diffusion-based methods.
    \item We semantically ground HOI generation by integrating a motion-text alignment loss to enforce consistency with language instructions, and conditioning on a hybrid 3D scene representation that encodes both geometric and semantic context from the surrounding scene.
    \item We design a reconstruction pipeline to extract large-scale, high-fidelity HOI data from egocentric videos, \rev{which we use to pretrain a grasping prior that improves grasp initialization and the grasp-to-manipulation transition}.
    \item On GRAB and HOT3D, FlowHOI attains the highest action-recognition accuracy and a 1.7$\times$ higher physics-simulation success rate over the strongest baseline. \rev{Retargeted to a real dexterous robot and executed open-loop, the generated interactions succeed in 57.5\% of trials on four contact-rich tasks, compared with 10\% for DiffH2O.}
\end{itemize}
\section{Related work}
\label{sec:related_work}

\noindent\textbf{Hand-Object Interaction Generation.}
Synthesizing articulated hand-object motions under realistic contact has been supported by both mocap- and vision-based datasets~\cite{kwon2021h2o,taheri2020grab,liu2022hoi4d,fan2023arctic,banerjee2025hot3d}.
Early kinematic methods leverage contact-aware priors, optimization, or implicit representations to reduce interpenetration~\cite{hassan2023synthesizing,grady2021contactopt,li2024task,karunratanakul2020grasping}, often with latent-variable models in canonical object spaces~\cite{jiang2021hand,taheri2022goal,zhang2021manipnet,zheng2023cams}.
Physics-based RL approaches ensure dynamic feasibility but scale poorly across objects and suffer from sim-to-real gaps~\cite{braun2024physically,rajeswaran2017learning}.
Recent diffusion methods~\cite{ghosh2023imos,liu2024geneoh,cha2024text2hoi} improve temporal coherence, yet DiffH2O~\cite{christen2024diffh2o} still suffers from physical artifacts, semantic inconsistencies, and slow inference, and LatentHOI~\cite{li2025latenthoi} remains limited in interaction length and data scale. Concurrent work~\cite{petrov2025tridi} extends diffusion to trilateral human-object-interaction modelling.
\rev{Object-centric generators synthesize 6-DoF object trajectories in 3D scenes from goal conditions~\cite{zeng2026gmt} or from egocentric observations with flow matching~\cite{saroha2026egoflow}, but do not model the articulated hand.}
We instead propose a flow matching framework that explicitly models distinct grasping and manipulation phases, achieving faster generation with improved physical plausibility and semantic alignment.

\noindent\textbf{Robot learning and motion priors from videos.}
Learning manipulation from human videos~\cite{mandlekar2019scaling,khazatsky2024droid, o2024open} is scalable but faces an embodiment gap. Egocentric datasets~\cite{banerjee2025hot3d, damen2020epic, perrett2025hd} partially bridge it, while VLA models~\cite{kim2024openvla,black2026pi0visionlanguageactionflowmodel, wang2025unified}, language-conditioned policies~\cite{ahn2022can, hao2024language}, and video generators~\cite{bharadhwaj2024gen2act,wang2025language,mao2025robot, goswami2025world} map perception to actions but lack explicit interaction structure. In parallel, diffusion-based human motion synthesis~\cite{tevet2022human,karunratanakul2023guided,chen2023executing,guo2022generating,petrovich2022temos,zhang2024motiondiffuse} and HOI-extending variants~\cite{xu2023interdiff,li2024controllable, peng2025hoi, wang2023physhoi} target full-body motion and rely on small mocap datasets~\cite{li2023object,bhatnagar22behave,lu2025humoto,mahmood2019amass}. Closely related to our reconstruction pipeline, WHOLE~\cite{ye2026whole} lifts hand-object interactions from egocentric videos into a world-grounded representation, but produces a per-clip reconstruction rather than a generative prior. \rev{World-space hand motion estimators~\cite{zhang2025hawor,zeng2026stablehand} likewise recover hand trajectories from egocentric videos, but estimate hand-only motion for each clip rather than learning a prior over hand-object interactions.} We instead acquire a robot-agnostic, fine-grained HOI sequence prior recovered from egocentric videos~\cite{hoque2025egodex} that can be retargeted to different embodiments.
\section{Preliminary: flow matching}
\label{sec:preliminary}

We briefly review the flow matching framework for generative modeling~\cite{lipman2022flow}.
Let $q(\mathbf{x})$ denote the unknown data distribution over $\mathbf{x}\in\mathbb{R}^d$ and $p_0(\mathbf{x})=\mathcal{N}(\mathbf{0},\mathbf{I})$ a simple prior. Flow matching learns a time-dependent vector field $\mathbf{v}(\mathbf{x},\tau)$ that transports samples from $p_0$ to a target distribution $p_1\approx q$ along a continuous probability path~\cite{chen2018neural}.
Since directly matching the marginal vector field of $p_\tau$ is generally intractable, \emph{conditional flow matching} (CFM)~\cite{lipman2022flow} constructs a tractable conditional path $p_\tau(\mathbf{x}\mid \mathbf{x}_1)$ for each data sample $\mathbf{x}_1\sim q$.
With the optimal-transport choice of linear interpolation,
\begin{equation}
\mathbf{x}_\tau = \bigl(1-(1-\sigma_{\min})\tau\bigr)\mathbf{x}_0 + \tau\,\mathbf{x}_1,
\quad \mathbf{x}_0\sim\mathcal{N}(\mathbf{0},\mathbf{I}),\ \tau\sim\mathcal{U}[0,1],
\label{eq:cfm_path}
\end{equation}
the conditional vector field admits a closed form:
\begin{equation}
\mathbf{u}_\tau(\mathbf{x}_\tau\mid \mathbf{x}_1)
= \frac{\mathbf{x}_1-(1-\sigma_{\min})\mathbf{x}_\tau}{1-(1-\sigma_{\min})\tau}.
\label{eq:cfm_target}
\end{equation}
A neural network $\mathbf{v}_\theta(\mathbf{x},\tau,\mathbf{c})$ conditioned on side information $\mathbf{c}$ is regressed to this target with a mean-squared loss. Samples are generated at inference by integrating $\mathbf{v}_\theta$ from $\tau\!=\!0$ to $\tau\!=\!1$ with a numerical ODE solver (we use $K$-step Euler). The role of $\sigma_{\min}$ and discretization details are deferred to Supp.~Sec.~\ref{sec:supp_implementation}.

In FlowHOI, instead of regressing the velocity $\mathbf{u}_\tau$, we adopt the equivalent $x$-prediction parameterization that directly regresses the clean target $\mathbf{x}_1$ for improved temporal stability~\cite{li2025back}. The corresponding analysis is in Supp.~Sec.~\ref{sec:supp_prediction_target}.

\section{Methodology}
\label{sec:method}
\begin{figure}[t]
    \centering
    \includegraphics[width=0.88\linewidth]{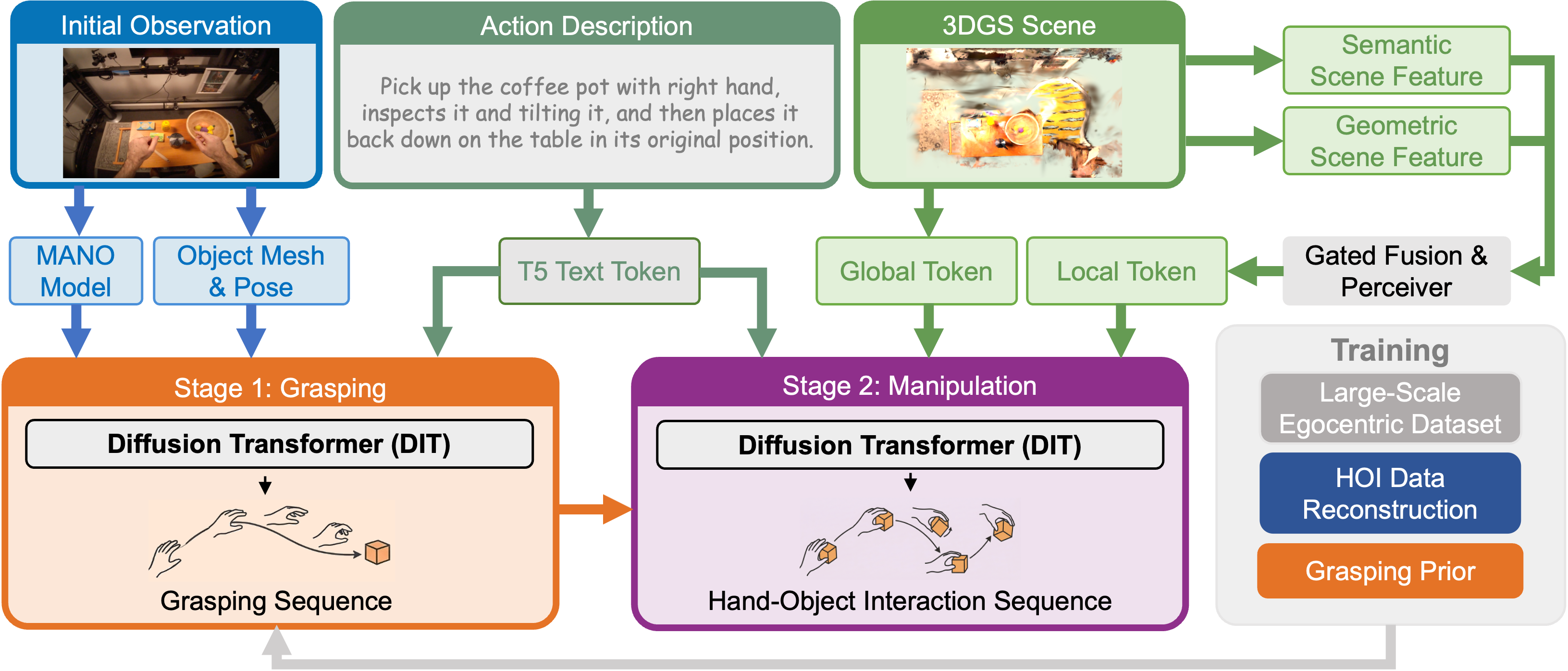}
    \caption{\textbf{Overview of FlowHOI.} Two-stage pipeline conditioned on an egocentric observation, text command, and 3D scene: (1) approach-and-grasp from a prior fine-tuned on reconstructed egocentric HOI data, and (2) manipulation conditioned on scene and language.}
    \label{fig:overview}
\end{figure}

In this section, we present our flow-based HOI motion generation conditioned on a single-frame initial egocentric observation, text command, and 3D scene context.
As illustrated in Fig.~\ref{fig:overview}, our framework consists of two stages: a \emph{Grasping} stage that generates approach-and-grasp motions from a pretrained prior fine-tuned on reconstructed egocentric HOI data, and a \emph{Manipulation} stage that produces subsequent interaction motions conditioned on 3D scene context and language instructions.

\subsection{Problem formulation}
\label{subsec:problem}

We address HOI motion generation in practical scenarios.
Given the first egocentric observation $I$, from which we extract the initial hand-object state $\mathbf{x}_{\text{init}}$ and object geometry $\mathcal{M}$, an action description $T$, and a 3DGS scene representation $\mathcal{G}$, our goal is to generate a temporally coherent hand-object interaction motion over $N$ frames.
We denote the object pose at frame $t$ as $\mathbf{O}_t$, and the left and right hand states as $\mathbf{H}^l_t$ and $\mathbf{H}^r_t$, respectively.
The full HOI trajectory is defined as
\begin{equation}
\mathbf{x} = \{(\mathbf{H}^l_t, \mathbf{H}^r_t, \mathbf{O}_t)\}_{t=0}^{N-1} \in \mathbb{R}^{N \times D},
\end{equation}
where $D$ denotes the per-frame feature dimension of the hand and object representations.
Given a dataset of $M$ paired interactions $\{(\mathbf{x}^{(i)}, \mathbf{c}^{(i)})\}_{i=1}^{M}$, our objective is to learn a conditional generative model $p_\theta(\mathbf{x}\mid \mathbf{c})$, where $\mathbf{c}$ is the conditioning information composed of encoded $\mathbf{x}_{\text{init}}$, $\mathcal{M}$, $T$, and $\mathcal{G}$.


\noindent\textbf{Data representation.}
Following~\cite{christen2024diffh2o}, each frame $t$ contains the object pose $\mathbf{O}_t=(\boldsymbol{p}^o_t, \boldsymbol{r}^o_t)\!\in\!\mathbb{R}^{9}$ with continuous 6D rotation~\cite{zhou2019continuity}, and bimanual MANO~\cite{MANO:SIGGRAPHASIA:2017} hand states $\mathbf{H}^h_t=(\tilde{\boldsymbol{p}}^h_t, \boldsymbol{r}^h_t, \boldsymbol{\theta}^h_t, \mathbf{s}^h_t)\!\in\!\mathbb{R}^{54}$ comprising root translation, 6D orientation, $24$D MANO PCA pose coefficients, and a per-joint signed-distance vector $\mathbf{s}^h_{t,k}=\boldsymbol{j}^{h,t}_k-\Pi_{\mathcal{M}_t}(\boldsymbol{j}^{h,t}_k)$ to the posed object surface that encourages contact-aware generation. Hand translations $\tilde{\boldsymbol{p}}^h_t=\boldsymbol{p}^h_t-\boldsymbol{p}^o_{t_g}$ are anchored at the transition-frame object position ($t_g\!=\!N_g\!-\!1$) to reduce inter-scene variance. Full notation is in Supp.~Sec.~\ref{sec:supp_implementation}.


\subsection{Hand-object data reconstruction from egocentric videos}
\label{subsec:data_reconstruction}
\begin{figure}[t]
    \centering
    \includegraphics[width=0.85\linewidth]{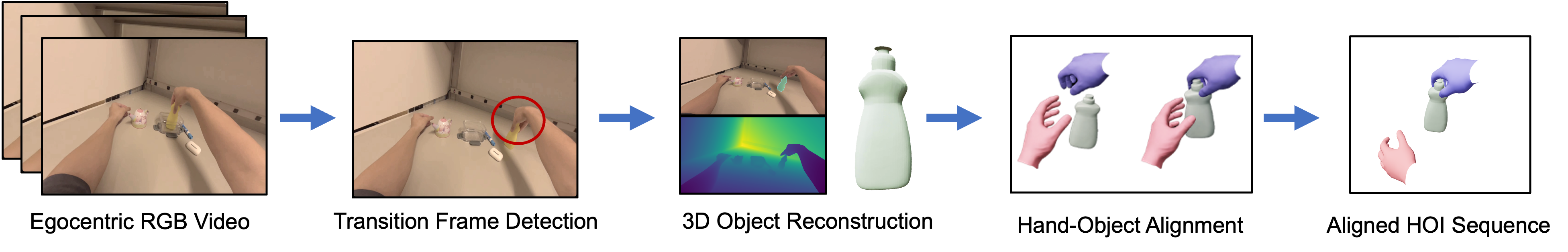}
    \caption{\textbf{Hand-object data reconstruction.} Detect the grasp-to-manipulation transition, reconstruct the object mesh via segmentation and metric depth, and align MANO under contact and non-penetration constraints. Detailed pipeline in supp.}
    \label{fig:data_reconstruction}
\end{figure}

Existing HOI datasets~\cite{taheri2020grab,chao2021dexycb} are orders of magnitude smaller than full-body MoCap collections~\cite{mahmood2019amass} due to severe self-occlusion and tightly coupled contact dynamics in egocentric views.
To bridge this gap, we build a reconstruction pipeline (Fig.~\ref{fig:data_reconstruction}) that converts raw egocentric videos from EgoDex~\cite{hoque2025egodex}, which provides RGB streams, camera parameters, text descriptions, and tracked hand keypoints but no object annotations, into high-fidelity HOI training data used exclusively to pretrain our grasping prior (Sec.~\ref{subsec:two_stage_generation}).

The pipeline consists of three steps.
\textbf{(1)~Transition detection:} we smooth wrist trajectories and identify the grasp-to-manipulation transition via local speed minima and direction changes.
\textbf{(2)~Object reconstruction:} we segment the target object with SAM3~\cite{carion2025sam}, estimate metric depth with DepthAnything3~\cite{lin2025depth}, and reconstruct a mesh with SAM3D~\cite{chen2025sam} from pre-transition frames where the object is static.
\textbf{(3)~Hand-object alignment:} we fit MANO meshes via inverse kinematics, optimize an object translation offset to satisfy fingerpad contact and non-penetration constraints at the transition frame, and propagate the alignment to all frames.
Full algorithmic details, loss formulations, hyperparameters, and evaluations of reconstruction results are provided in the supplementary material.

\subsection{Two-stage hand-object interaction generation}
\label{subsec:two_stage_generation}
Motivated by the human grasp-then-manipulate motor pattern~\cite{johansson2009coding}, we decompose HOI into a \emph{Grasping} and a \emph{Manipulation} phase. While DiffH2O~\cite{christen2024diffh2o} also separates these phases, our key design is \emph{stage-specific conditioning}: grasping uses only object geometry, language, and the initial state, while manipulation additionally consumes 3DGS scene tokens with a motion--text alignment objective, so each stage focuses on its most relevant signals---geometry/reachability vs.\ semantics/scene---rather than entangling all sources in a single denoiser.

\noindent\textbf{Initial state estimation.}
From the first egocentric observation, we estimate the initial hand pose with~\cite{potamias2025wilor}, recover the object mesh and 6D pose with SAM3D~\cite{chen2025sam}, and lift both into a common world frame defined by the first camera's extrinsics, yielding $\mathbf{x}_{\text{init}}=(\mathbf{O}_0,\mathbf{H}^l_0,\mathbf{H}^r_0)$. We encode object geometry $\mathcal{M}$ with a Basis Point Set~\cite{prokudin2019efficient} and the action description $T$ with a frozen T5-Large~\cite{raffel2020exploring}; both are reused at every Euler step.

\noindent\textbf{Hybrid 3D scene tokens.}
Manipulation is constrained by both coarse spatial layout and fine-grained, language-aligned object semantics, so we build complementary \emph{global} and \emph{local} tokens from a single 3DGS reconstruction of the scene~\cite{gu2024egolifter, lv2025photoreal} (with moving hands and objects masked out). We sample $N_s$ Gaussian centroids via Farthest Point Sampling~\cite{eldar1997farthest}, $\mathbf{X}\!\in\!\mathbb{R}^{N_s\times 3}$, and extract a geometric embedding via Concerto~\cite{zhang2025concerto} and a language-aligned semantic embedding via SceneSplat~\cite{li2025scenesplat}, stacked into $\mathbf{E}\!\in\!\mathbb{R}^{N_s\times d_e}$ and $\mathbf{U}\!\in\!\mathbb{R}^{N_s\times d_u}$.

We fuse semantic and geometric cues with a linear-complexity gated mechanism adapted from squeeze-and-excitation~\cite{hu2018squeeze}: after projecting both modalities into a shared $d_h$-dim space, $\tilde{\mathbf{E}}\!=\!\Phi_e(\mathbf{E})$ and $\tilde{\mathbf{U}}\!=\!\Phi_u(\mathbf{U})$, we combine them with a learnable channel-wise gate $\boldsymbol{\alpha}\!\in\!\mathbb{R}^{d_h}$:
\begin{equation}
\mathbf{F} = \tilde{\mathbf{E}} + \sigma(\boldsymbol{\alpha}) \odot \tilde{\mathbf{U}},
\end{equation}
allowing adaptive modulation of semantics and geometry while preserving linear complexity.

To explicitly encode spatial layout, we concatenate a Fourier positional encoding $\gamma(\mathbf{X}) \in \mathbb{R}^{N_s \times d_h}$ of the 3D point coordinates and apply a final projection to get the hybrid scene tokens:
\begin{equation}
\mathbf{P} = \Psi\big([\mathbf{F}, \gamma(\mathbf{X})]\big) \in \mathbb{R}^{N_s \times d_s},
\end{equation}
where $d_s$ denotes the dimension of the scene token.
Cross-attending each of the $N$ per-frame motion tokens to all $N_s$ scene tokens would cost $\mathcal{O}(N\!\cdot\!N_s)$, so we compress $\mathbf{P}$ into $L\!\ll\!N_s$ latents via a Perceiver bottleneck~\cite{jaegle2021perceiver},
\begin{equation}
\mathbf{S}_{\text{local}} = \mathrm{Perceiver}(\mathbf{P}; \mathbf{Q}_0) \in \mathbb{R}^{L \times d_s},
\end{equation}
with learnable queries $\mathbf{Q}_0\!\in\!\mathbb{R}^{L\times d_s}$, reducing the per-layer cost to $\mathcal{O}(L\!\cdot\!N_s + N\!\cdot\!L)$.

For coarse layout, we voxelize the scene into an occupancy grid $\mathbf{V}\!\in\!\{0,1\}^{H_z\times H_x\times H_y}$, split it into non-overlapping $p\!\times\!p\!\times\!p$ cubic patches, prepend a learnable [\textsc{cls}] token, and feed the sequence into a 3D vision transformer~\cite{jiang2024scaling}; its [\textsc{cls}] output is the global token $\mathbf{S}_{\text{global}} = \mathrm{ViT}(\mathbf{V}) \in \mathbb{R}^{1 \times d_s}$. Hyperparameters ($H_x,H_y,H_z$, $p$, $N_s$, $L$) are in the supplementary. The two representations play complementary roles: \rev{$\mathbf{S}_{\text{global}}$ mainly supports scene-aware interaction placement and the selection of the interacting hand, while $\mathbf{S}_{\text{local}}$ refines fine-grained hand-object contact geometry tied to specific interaction regions (Supp.~Sec.~\ref{sec:supp_scene_granularity}).}

\noindent\textbf{Grasping stage.}
Since the object is static during grasping, we generate only the hand state $\mathbf{x}^g\!=\!(\mathbf{H}^l,\mathbf{H}^r)\!\in\!\mathbb{R}^{N_g\times D_h}$ ($D_h$: per-frame hand feature dim), conditioned on
\begin{equation}
\mathbf{c}^g = \{\, \mathrm{BPS}(\mathcal{M});\ \mathrm{T5}(T_g);\ \mathbf{x}_{\text{init}} \,\},
\end{equation}
where $T_g$ is a grasp-focused sub-instruction extracted from $T$ via an MLLM~\cite{achiam2023gpt}, isolating grasping conditioning from manipulation semantics, while $\mathbf{x}_{\text{init}}$ is concatenated to the noisy motion token at every sampling step to anchor generation.
We train a network $f^g_\theta$ with $x$-prediction~\cite{li2025back} to recover the clean target $\mathbf{x}_1^g$ from $\mathbf{x}_\tau^g$ along the linear path of Eq.~\eqref{eq:cfm_path},
\begin{equation}
\mathcal{L}_{\text{flow}}^g
= \mathbb{E}_{\tau,\mathbf{x}_0,\mathbf{x}_1^g}
\bigl\|f^g_\theta(\mathbf{x}_\tau^g, \tau, \mathbf{c}^g) - \mathbf{x}_1^g\bigr\|_2^2,
\end{equation}
with the corresponding velocity field recovered through Eq.~\eqref{eq:cfm_target}.
To ground generation in language semantics, we add a TMR-inspired symmetric InfoNCE alignment loss~\cite{petrovich2023tmr,oord2018representation}
\begin{equation}
\label{eq:align}
\mathcal{L}_{\text{align}} = \tfrac{1}{2}(\mathcal{L}_{\text{t2m}} + \mathcal{L}_{\text{m2t}}),
\end{equation}
that aligns text and motion embeddings in a shared latent space.
The total grasping loss is $\mathcal{L}_{\text{grasp}} = \mathcal{L}^g_{\text{flow}} + \lambda_{\text{align}}\,\mathcal{L}_{\text{align}}$ with $\lambda_{\text{align}}\!=\!0.1$. At inference we Euler-integrate the recovered vector field from $\tau\!=\!0$ to $\tau\!=\!1$ to obtain the grasping trajectory terminating at $t_g$.


\noindent\textbf{Manipulation stage.}
Unlike grasping, manipulation requires reasoning over the full sequence $\mathbf{x}^m\!\in\!\mathbb{R}^{N\times D}$ while preserving the established grasp and respecting scene constraints. We treat the previously generated grasping trajectory $\mathbf{x}^g_{[0,t_g]}$ as a known prefix and condition the manipulation generator on
\begin{equation}
\mathbf{c}^{m} = \{\, \mathrm{BPS}(\mathcal{M});\ \mathrm{T5}(T);\ \mathbf{S}_{\text{local}};\ \mathbf{S}_{\text{global}};\ \mathbf{x}^{g}_{t_g}\,\}.
\end{equation}
With temporal mask $\mathbf{M}\!\in\!\{0,1\}^{N}$ (zero on $t\!\le\!t_g$, one elsewhere), at each Euler step with flow time $\tau\!<\!0.9$ we softly inpaint the grasping prefix into the noisy trajectory,
\begin{equation}
\bar{\mathbf{x}}^m_\tau = \mathbf{M}\odot \mathbf{x}^m_\tau + (1-\mathbf{M})\odot \mathbf{x}^g_{[0,t_g]},
\end{equation}
where $\mathbf{x}^g_{[0,t_g]}$ is padded with the static initial object pose $\mathbf{O}_0$ to match the dimension of $\mathbf{x}^m$. Disabling the replacement for $\tau\!\geq\!0.9$ lets the model refine the full sequence, and we clamp the transition via $\bar{\mathbf{x}}^m_\tau[t_g]\!\leftarrow\!\mathbf{x}^g_{t_g}$. Compared to DiffH2O's diffusion-step inpainting~\cite{christen2024diffh2o}, our continuous-time formulation enforces transition consistency at the ODE level.
Training uses a masked $x_1$-prediction loss applied only to the manipulation portion,
\begin{equation}
\mathcal{L}_{\text{flow}}^m
= \mathbb{E}_{\tau,\mathbf{x}_0,\mathbf{x}_1^m}
\bigl\|\mathbf{M}\!\odot\!\bigl(f^m_\theta(\bar{\mathbf{x}}^m_\tau,\tau,\mathbf{c}^m) - \mathbf{x}^m_1\bigr)\bigr\|_2^2,
\end{equation}
and the total manipulation loss is $\mathcal{L}_{\text{manip}} = \mathcal{L}_{\text{flow}}^m + \lambda_{\text{align}}\,\mathcal{L}_{\text{align}}$ (Eq.~\ref{eq:align}).


\section{Experiments}
\label{sec:exp}

\subsection{Datasets}
\label{subsec:datasets}
We evaluate our method on two datasets, namely HOT3D~\cite{banerjee2025hot3d} and GRAB~\cite{taheri2020grab}. 
We pretrain our grasping model with our curated EgoDex~\cite{hoque2025egodex} dataset as described in Sec.~\ref{subsec:data_reconstruction}. EgoDex is used solely for training the grasping prior model and excluded from our evaluations.
GRAB~\cite{taheri2020grab} provides high-fidelity hand-object motion capture and is used for quantitative evaluation without scene context. HOT3D~\cite{banerjee2025hot3d} offers real-world egocentric recordings with accurate hand and object pose annotations together with the source video required by our 3DGS scene reconstruction, enabling scene-conditioned evaluation and generalization to unseen objects. HOT3D's annotations are noisier than GRAB's mocap supervision: even Real Mocap reaches only 6.67\% physics-simulation SR (Table~\ref{tab:main_results}), so we treat CR and AR as the more informative signals on HOT3D (analysis in Supp.~Sec.~\ref{sec:supp_dataset}).
Note that neither dataset provides natural language action descriptions or pre-built 3D scene reconstructions. Details of our reconstruction and annotation procedures are provided in the supplementary material.

\subsection{Implementation details}
\label{subsec:implementation_details}
We implement FlowHOI in PyTorch~\cite{paszke2019pytorch} and train on an RTX 6000 Ada GPU (48GB) with batch size 64, using conditional flow matching~\cite{lipman2022flow} with optimal-transport paths, $x$-prediction, and 50-step Euler integration at inference. For DiffH2O~\cite{christen2024diffh2o} and LatentHOI~\cite{li2025latenthoi}, we run the official implementations and retrain on GRAB and HOT3D under the same data splits and evaluation protocol. For fair comparison we replace the CLIP~\cite{radford2021learning} text encoder with the more capable T5~\cite{raffel2020exploring} encoder for all baselines. Additional details are in the supplementary material.
\begin{figure}[t]
    \centering
    \includegraphics[width=0.95\linewidth]{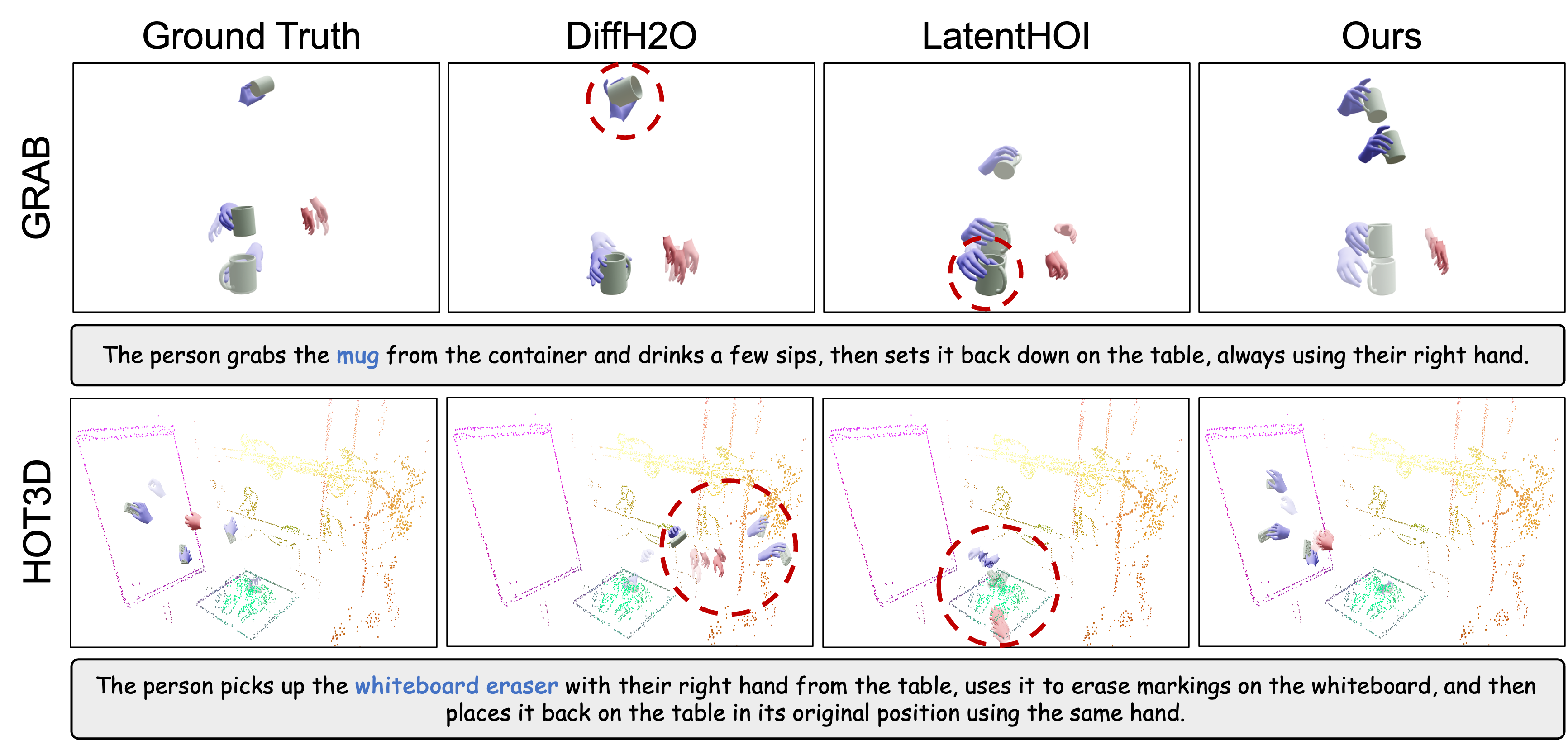}
    \caption{Qualitative comparison vs.\ DiffH2O~\cite{christen2024diffh2o} and LatentHOI~\cite{li2025latenthoi} (top: GRAB, bottom: HOT3D with 3D scene). Ours yields more natural grasps and physically plausible manipulations that better follow the instruction. Best viewed in the supplementary video.}
    \label{fig:qualitative}
\end{figure}

\begin{table}[t!]
    \caption{Quantitative evaluation on GRAB and HOT3D. Best among generation methods in \textbf{bold}; Real Mocap is the upper bound. IV/ID are bolded only when paired with non-trivial CR (cf.\ LatentHOI on HOT3D: low IV/ID with $0\%$ physics SR; Supp.~Sec.~\ref{sec:supp_dataset}). Binomial SE on SR $\le\!1.1\%$ on GRAB and $\le\!0.4\%$ on HOT3D.}
  \label{tab:main_results}
  \centering
  \resizebox{\textwidth}{!}{%
    \begin{tabular}{@{}c|l|ccccccccccc@{}}
      \toprule
      & & \multicolumn{4}{c}{Physical Interaction Quality} & \multicolumn{3}{c}{Motion Quality} & \multicolumn{4}{c}{Realizable Physical Feasibility} \\
      \cmidrule(lr){3-6} \cmidrule(lr){7-9} \cmidrule(lr){10-13}
      Dataset & Method
      & IV [$cm^3$] ($\downarrow$) 
      & ID [$cm$] ($\downarrow$) 
      & CR [\%] ($\uparrow$) 
      & IVU ($\downarrow$) 
      & AR ($\uparrow$) 
      & SD [$m$] ($\uparrow$) 
      & OD [$m$] ($\uparrow$) 
      & Phy [\%] ($\uparrow$) 
      & SR [\%] ($\uparrow$) 
      & HT [s] ($\uparrow$) 
      & Time [s] ($\downarrow$) 
      \\
      \midrule
      \multirow{4}{*}{GRAB}
      & Real Mocap          &  6.49                       & 0.43                       &  6.62                       & 0.09                       & 1.00                       & -                          & 0.18                       & 95.56 & 77.06    & 2.40    & -    \\
      & DiffH2O             & 13.11                       & 1.32                       &  8.14                       & 0.16                       & 0.87                       & 0.12                          & 0.13              & 89.44 & 33.03    & 0.35    & 6.34 \\
      & LatentHOI       & 13.76                       & \textbf{0.99}              &  \textbf{9.18}              & 0.14                       & 0.78                          & 0.11                          & 0.14                       & \textbf{96.40} & 28.44    & 0.29    & 3.57 \\      

      & Ours                & \textbf{10.93}              & 1.28                       &  6.85                       & \textbf{0.13}              & \textbf{0.95}              & \textbf{0.13}                 & \textbf{0.16}              & 90.90 & \textbf{55.96}    & \textbf{1.50}    & \textbf{0.16} \\
      \midrule
      \multirow{4}{*}{HOT3D}
      & Real Mocap   & 2.88 & 0.63 & 1.69 & 0.15 & 1.00 & - & 0.23 & 16.80 & 6.67 & 0.08 & - \\
      & DiffH2O & 3.25 & 0.66 & 2.07 & \textbf{0.18} & 0.71 & \textbf{0.15} & \textbf{0.23} & 16.47 & 4.00 & 0.03 & 6.54 \\
      & LatentHOI           & \textbf{1.54} & \textbf{0.44} & 1.59 & 0.21 & 0.65 & 0.12 & 0.18 & 14.36 & 0.00 & 0.00 & 3.21 \\
      & Ours                & 3.36 & 0.72 & \textbf{2.14} & 0.19 & \textbf{0.78} & 0.10 & 0.20 & \textbf{17.15} & \textbf{5.33} & \textbf{0.03} & \textbf{0.16} \\
      \bottomrule
    \end{tabular}
  }
\end{table}


\subsection{Evaluation metrics}
\label{subsec:metrics}
Aligned with prior work~\cite{christen2024diffh2o,li2025latenthoi,ghosh2023imos,tendulkar2023flex}, we report three metric groups (formal definitions in Supp.~Sec.~\ref{sec:supp_metrics}). \emph{Physical interaction quality} comprises interpenetration volume (IV), depth (ID), contact ratio (CR), and IV per contact unit (IVU). \emph{Motion quality} comprises action-recognition accuracy (AR) and sample/overall diversity (SD/OD). \emph{Realizable physical feasibility} comprises two complementary measures. The heuristic physical plausibility score (Phy) of~\cite{li2025latenthoi} is a kinematic-only check (sustained contact and object above ground) and does not enforce dynamics. For physics-grounded evaluation, we additionally retarget each generated HOI to the Allegro Hand via IK~\cite{kim2025pyroki} and execute it in Isaac Gym~\cite{makoviychuk2021isaac} with a physics-based tracker~\cite{liu2025dextrack}, reporting the simulation success rate (SR), holding time (HT), and the per-sequence HOI generation time. We treat SR as the primary holistic indicator, since low IV/IVU can be trivially achieved by avoiding contact.

\subsection{HOI generation comparison}
We compare FlowHOI with DiffH2O~\cite{christen2024diffh2o} and LatentHOI~\cite{li2025latenthoi} on GRAB and HOT3D (Table~\ref{tab:main_results}), focusing on long-horizon contact consistency, semantic alignment, and inference efficiency.

\noindent\textbf{Contact consistency over long horizons.}
On GRAB, FlowHOI achieves the lowest IV and IVU, indicating that contact geometry remains consistent over extended interactions with less error accumulation.
On HOT3D, FlowHOI attains the highest contact ratio while remaining competitive with DiffH2O on penetration metrics. LatentHOI achieves lower IV/ID at the cost of substantially weaker contact and a 0\% physics-simulation SR (Table~\ref{tab:main_results}); a detailed analysis of its reproduction is in Supp.~Sec.~\ref{sec:supp_dataset}.

\noindent\textbf{Semantically grounded generation without sacrificing diversity.}
FlowHOI achieves the highest AR, reflecting strong language and scene grounding from the manipulation stage's 3D scene tokens and motion-text alignment loss. Sample-level and overall diversity remain comparable to prior methods, with the highest OD on GRAB.

\noindent\textbf{Realizable physical feasibility and efficiency.}
In physics simulation, FlowHOI achieves the best SR on GRAB (55.96\% vs.\ 28.44\% for LatentHOI and 33.03\% for DiffH2O) and the longest execution duration of 1.5\,s, confirming that the generated trajectories remain stable and executable after retargeting. Although LatentHOI achieves the highest heuristic Phy score on GRAB, this metric does not capture physical stability under dynamics, as reflected by its substantially lower SR.
By adopting flow matching, the planning step requires only $0.16$\,s per sequence, a $40\times$ reduction relative to the diffusion-step ODE of DiffH2O ($6.34$\,s) and LatentHOI ($3.57$\,s). 
Qualitative results in Fig.~\ref{fig:qualitative} further corroborate these findings, showing more realistic grasp configurations and smoother manipulation trajectories, particularly for long-horizon interactions.

\rev{\noindent\textbf{Input-matched comparison.}
No method uses scene context on GRAB, so the comparison there is input-matched.
On HOT3D, FlowHOI additionally receives 3D scene tokens that the baselines do not use.
Without scene conditioning, FlowHOI reaches an AR of 0.73 on HOT3D (Table~\ref{tab:ablation_scene}), compared with 0.71 for DiffH2O and 0.65 for LatentHOI.
The full HOT3D results in Table~\ref{tab:main_results} thus combine the generator with scene conditioning, whose contribution Table~\ref{tab:ablation_scene} and Supp.~Sec.~\ref{sec:supp_scene_granularity} isolate.}

\rev{\noindent\textbf{Post-grasp motion.}
After the transition frame, the manipulation stage continues to predict finger articulation and contact states, so the object can rotate relative to the hand instead of being carried rigidly.
On 174 GRAB hand instances with an asymmetric object and more than $30^\circ$ of ground-truth hand-relative reorientation, FlowHOI reduces the hand-relative rotation error from $83.0^\circ$ for DiffH2O to $46.4^\circ$ (Supp.~Sec.~\ref{sec:supp_dexterity}).}

\subsection{Showcase of real-world applications}
\label{subsec:real_world}
\begin{figure}[t]
    \centering
    \includegraphics[width=0.85\linewidth]{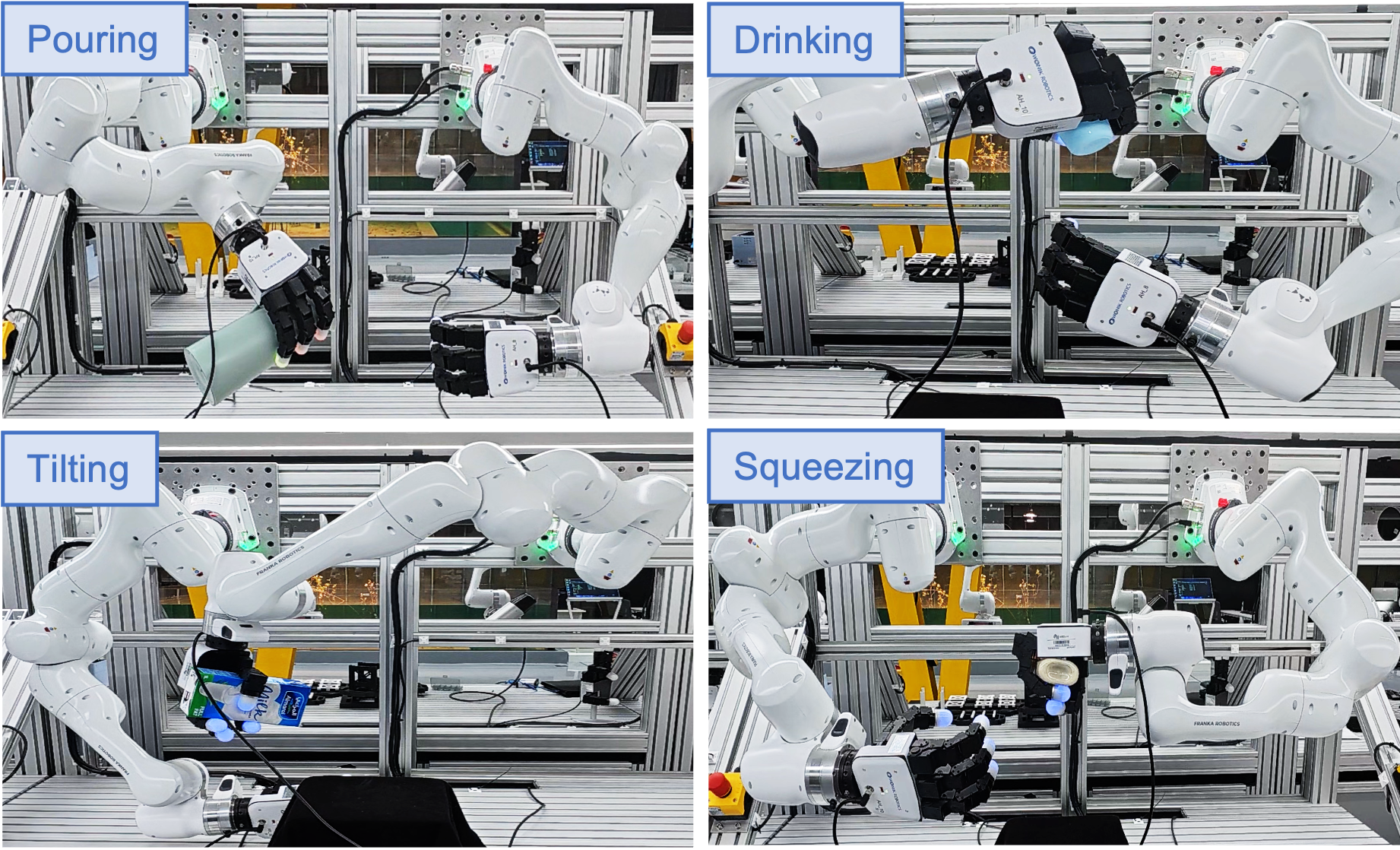}
    \caption{Real-world execution on four contact-rich tasks (pouring, drinking, tilting, squeezing): generated HOI retargeted to a Franka Panda + Allegro Hand and run open-loop.}
    \label{fig:real_world}
\end{figure}

We further evaluate the physical feasibility of executing generated HOI sequences on a real-world dexterous manipulation platform, consisting of two Franka Emika Panda robotic arms~\cite{haddadin2024franka}, each equipped with an Allegro Hand v5~\cite{allegrohand}. We consider four contact-rich household manipulation tasks: drinking from a cup, pouring liquid between containers of different sizes, tilting a container, and squeezing dressing. In this setup FlowHOI plays the role of an interaction \emph{planner} rather than a closed-loop controller: it produces a single reference trajectory from one observation, which is then refined offline and executed open-loop.
The pipeline takes egocentric RGB and a Gaussian-LIC~\cite{lang2025gaussian} 3DGS scene as input (Fig.~\ref{fig:teaser}). The pyroki solver~\cite{kim2025pyroki} converts the robot proprioceptive state into the initial MANO pose required by FlowHOI; after generation, the same solver retargets the predicted MANO trajectory to 16-DOF Allegro joint actions, which are refined by DexTrack~\cite{liu2025dextrack} and executed open-loop via a joint impedance controller, producing stable contact-rich interactions (Fig.~\ref{fig:real_world}).
Quantitatively (Table~\ref{tab:realworld}), FlowHOI achieves a 57.5\% average SR, $5.75\times$ DiffH2O's 10\%. Residual failures stem primarily from embodiment mismatch (e.g., Allegro fingertips not fitting the mug handle for pouring) and post-grasp contact slippage under open-loop tracking.
\rev{Table~\ref{tab:realworld} therefore measures executability under nominal conditions, not robustness to state-estimation errors or disturbances.}

%
\begin{table}[t]
  \caption{Real-world execution success rate (SR, out of 10 trials) on four contact-rich tasks. Aggregated over 40 trials, FlowHOI achieves $57.5\%$ [$42.2\%, 71.5\%$] vs.\ DiffH2O $10.0\%$ [$4.0\%, 23.0\%$] ($95\%$ Wilson CI); the intervals do not overlap.}
  \label{tab:realworld}
  \centering
  \footnotesize
  \begin{tabular}{@{}l|cccc|c@{}}
    \toprule
    Method & Pouring & Drinking & Tilting & Squeezing & Avg.\ \\
    \midrule
    DiffH2O~\cite{christen2024diffh2o}   & 2/10          & 0/10          & 2/10          & 0/10          & 10.0\% \\
    FlowHOI (Ours)                       & \textbf{5/10} & \textbf{4/10} & \textbf{7/10} & \textbf{7/10} & \textbf{57.5\%} \\
    \bottomrule
  \end{tabular}
\end{table}%

\subsection{Ablation study}


\rev{\noindent\textbf{Two-stage decomposition.}
We compare with a single-stage FlowHOI that generates the full sequence in one pass and is matched in representation, training data, objective, and parameter budget (Supp.~Sec.~\ref{sec:supp_single_stage}).
The two-stage design raises the contact ratio and Phy and lowers interpenetration on both GRAB and HOT3D.
AR and simulation SR are comparable, and the single-stage model is 40\,ms faster.
The benefit of the decomposition thus lies in contact establishment rather than overall accuracy, and it also enables stage-specific conditioning and grasp-only pretraining.}

\noindent\textbf{Effect of Pretraining Grasping Model.}
Table~\ref{tab:ablation_prior} compares grasping models with and without pretraining on large-scale egocentric data (Sec.~\ref{subsec:data_reconstruction}). We report grasp error (GE), defined as the distance between the generated hand pose and the ground-truth grasp at the end of the grasping stage. Pretraining substantially reduces GE (0.10\,m\,$\to$\,0.06\,m, 40\% relative reduction), indicating more accurate grasp initialization. Penetration-related metrics remain comparable, indicating the prior improves grasp accuracy without regressing the contact behavior already enforced by our contact-aware data representation.
\rev{With the manipulation generator held fixed and only the grasp prefix varied (Supp.~Sec.~\ref{sec:supp_egodex}), pretraining also lowers the wrist velocity across the stage boundary from 0.583 to 0.340\,m/s and raises handedness accuracy from 96.34\% to 99.02\%, while Phy and simulation SR show no clear difference at our evaluation scale.}


\noindent\textbf{Semantic grounding via text and scene conditioning.}
Results are summarized in Table~\ref{tab:ablation_text_scene}. For text encoding, the T5 encoder consistently outperforms CLIP-based variants in action recognition accuracy.
Incorporating the motion-text alignment loss (Eq.~\eqref{eq:align}) further improves the performance, indicating stronger correspondence between generated motions and language instructions.
For scene encoding \rev{on HOT3D}, removing scene information leads to degraded action recognition accuracy and large final displacement error (FDE). Geometry-only or semantics-only scene representations provide partial improvements, while our fused representation achieves the best performance on both metrics. This demonstrates that jointly modeling spatial geometric constraints and scene semantics is critical for accurate object motion and semantically consistent manipulation.
\begin{table}[t]
  \centering
  \caption{Ablation studies. (a) Scene encoder choices \rev{on HOT3D}; FDE: final displacement error of object pose. (b) Effect of pretraining with large-scale egocentric HOI data on GRAB; GE: grasp error at the end of the grasping stage. (c) Text encoder choices.}
  \label{tab:ablation_text_scene}

  \begin{subtable}[c]{0.30\linewidth}
    \centering
    \caption{Scene encoder.}
    \label{tab:ablation_scene}
    \resizebox{\linewidth}{!}{%
      \begin{tabular}{@{}l|cc@{}}
        \toprule
        Method & AR ($\uparrow$) & FDE ($\downarrow$) \\
        \midrule
        w/o scene                            & 0.73          & 0.33 \\
        PointNet++~\cite{qi2017pointnet++}   & 0.59          & 0.44 \\
        Concerto~\cite{zhang2025concerto}    & 0.75          & 0.26 \\
        SceneSplat~\cite{li2025scenesplat}   & 0.73          & 0.30 \\
        Ours                                 & \textbf{0.78} & \textbf{0.24} \\
        \bottomrule
      \end{tabular}
    }
  \end{subtable}
  \hfill
  \begin{subtable}[c]{0.46\linewidth}
    \centering
    \caption{Pretraining grasping prior on GRAB.}
    \label{tab:ablation_prior}
    \resizebox{\linewidth}{!}{%
      \begin{tabular}{@{}l|ccccc@{}}
        \toprule
        Model
        & IV [$cm^3$] ($\downarrow$)
        & ID [$cm$] ($\downarrow$)
        & CR [\%] ($\uparrow$)
        & IVU ($\downarrow$)
        & GE [$m$] ($\downarrow$) \\
        \midrule
        w/o prior   & \textbf{10.40} & 1.31          & 6.78          & 0.13 & 0.10 \\
        w/ prior    & 10.93          & \textbf{1.28} & \textbf{6.85} & 0.13 & \textbf{0.06} \\
        \bottomrule
      \end{tabular}
    }
  \end{subtable}
  \hfill
  \begin{subtable}[c]{0.20\linewidth}
    \centering
    \caption{Text encoder.}
    \label{tab:ablation_text}
    \resizebox{\linewidth}{!}{%
      \begin{tabular}{@{}l|c@{}}
        \toprule
        Method & AR ($\uparrow$) \\
        \midrule
        CLIP~\cite{radford2021learning}             & 0.89 \\
        CLIP + Align. Loss                          & 0.88 \\
        T5~\cite{raffel2020exploring}               & 0.92 \\
        T5 + Align. Loss                            & \textbf{0.94} \\
        \bottomrule
      \end{tabular}
    }
  \end{subtable}
\end{table}


\section{Conclusion}
\label{sec:conclusion}

We presented FlowHOI, a two-stage flow-matching framework that generates semantically grounded HOI sequences conditioned on an egocentric observation, a language instruction, and a 3DGS scene reconstruction. By decoupling geometry-centric grasping from semantics-centric manipulation with 3D scene tokens and a motion-text alignment loss, FlowHOI grounds interactions in both the physical scene and the language instruction. A reconstruction pipeline recovering HOI trajectories from egocentric videos further provides \rev{a grasping prior that improves grasp initialization}. On GRAB and HOT3D, FlowHOI achieves the highest action-recognition accuracy, a 1.7$\times$ higher physics-simulation success rate and a 40$\times$ inference speedup over DiffH2O~\cite{christen2024diffh2o} (the strongest diffusion-based baseline in our evaluation), and up to ${\sim}21\%$ lower interpenetration on GRAB compared to LatentHOI~\cite{li2025latenthoi}. \rev{Real-robot experiments on four contact-rich tasks show that the generated interactions can be retargeted to a dexterous robot hand and executed open-loop.}

\noindent\textbf{Limitations and future work.} \rev{FlowHOI targets contact-rich articulated-hand HOI generation and retargeting under nominal conditions, and within this scope it has three limitations.
(i)~It relies on initial-state estimation from a single observation.
(ii)~It generates purely kinematic trajectories, which on the robot depend on offline refinement by a physics-based tracker~\cite{liu2025dextrack} and are executed open-loop, so FlowHOI cannot react online to perturbations or contact slippage.
(iii)~Each sequence contains a single grasp-to-manipulation transition. The generated post-grasp motion includes hand-relative reorientation, but we do not evaluate repeated release and regrasp, force-level contact modes, deformable objects, or general in-hand manipulation, and multi-phase tasks such as opening a lid are not covered.} Future work includes closed-loop \rev{state estimation, replanning, and contact recovery, a phase planner that chains the grasping and manipulation generators across multi-segment tasks}, scaling to mobile manipulation by jointly modeling base motion and reachability with HOI generation, and learning priors from large-scale exocentric videos.



\begin{ack}
This work is partially supported by ``The MBZUAI-WIS Joint Program for Artificial Intelligence Research''.
We thank the Agentic Robotics Lab at MBZUAI for their hardware support.
\end{ack}

\clearpage
\bibliographystyle{unsrtnat}
\bibliography{references}

@inproceedings{peebles2023scalable,
  title={Scalable diffusion models with transformers},
  author={Peebles, William and Xie, Saining},
  booktitle={Proceedings of the IEEE/CVF international conference on computer vision},
  pages={4195--4205},
  year={2023}
}

@article{raffel2020exploring,
  title={Exploring the limits of transfer learning with a unified text-to-text transformer},
  author={Raffel, Colin and Shazeer, Noam and Roberts, Adam and Lee, Katherine and Narang, Sharan and Matena, Michael and Zhou, Yanqi and Li, Wei and Liu, Peter J},
  journal={Journal of machine learning research},
  volume={21},
  number={140},
  pages={1--67},
  year={2020}
}

@article{zhang2025concerto,
  title={Concerto: Joint 2d-3d self-supervised learning emerges spatial representations},
  author={Zhang, Yujia and Wu, Xiaoyang and Lao, Yixing and Wang, Chengyao and Tian, Zhuotao and Wang, Naiyan and Zhao, Hengshuang},
  journal={arXiv preprint arXiv:2510.23607},
  year={2025}
}

@inproceedings{prokudin2019efficient,
  title={Efficient learning on point clouds with basis point sets},
  author={Prokudin, Sergey and Lassner, Christoph and Romero, Javier},
  booktitle={Proceedings of the IEEE/CVF international conference on computer vision},
  pages={4332--4341},
  year={2019}
}

@inproceedings{jaegle2021perceiver,
  title={Perceiver: General perception with iterative attention},
  author={Jaegle, Andrew and Gimeno, Felix and Brock, Andy and Vinyals, Oriol and Zisserman, Andrew and Carreira, Joao},
  booktitle={International conference on machine learning},
  pages={4651--4664},
  year={2021},
  organization={PMLR}
}

@article{loshchilov2017decoupled,
  title={Decoupled weight decay regularization},
  author={Loshchilov, Ilya and Hutter, Frank},
  journal={arXiv preprint arXiv:1711.05101},
  year={2017}
}

@article{li2025back,
  title={Back to basics: Let denoising generative models denoise},
  author={Li, Tianhong and He, Kaiming},
  journal={arXiv preprint arXiv:2511.13720},
  year={2025}
}

@inproceedings{taheri2020grab,
  title={GRAB: A dataset of whole-body human grasping of objects},
  author={Taheri, Omid and Ghorbani, Nima and Black, Michael J and Tzionas, Dimitrios},
  booktitle={European conference on computer vision},
  pages={581--600},
  year={2020},
  organization={Springer}
}

@article{flash1985coordination,
  title={The coordination of arm movements: an experimentally confirmed mathematical model},
  author={Flash, Tamar and Hogan, Neville},
  journal={Journal of neuroscience},
  volume={5},
  number={7},
  pages={1688--1703},
  year={1985},
  publisher={Society for Neuroscience}
}

@article{chung2020gaussian,
  title={Gaussian kernel smoothing},
  author={Chung, Moo K},
  journal={arXiv preprint arXiv:2007.09539},
  year={2020}
}

@article{carion2025sam,
  title={Sam 3: Segment anything with concepts},
  author={Carion, Nicolas and Gustafson, Laura and Hu, Yuan-Ting and Debnath, Shoubhik and Hu, Ronghang and Suris, Didac and Ryali, Chaitanya and Alwala, Kalyan Vasudev and Khedr, Haitham and Huang, Andrew and others},
  journal={arXiv preprint arXiv:2511.16719},
  year={2025}
}

@article{lin2025depth,
  title={Depth anything 3: Recovering the visual space from any views},
  author={Lin, Haotong and Chen, Sili and Liew, Junhao and Chen, Donny Y and Li, Zhenyu and Shi, Guang and Feng, Jiashi and Kang, Bingyi},
  journal={arXiv preprint arXiv:2511.10647},
  year={2025}
}

@article{chen2025sam,
  title={Sam 3d: 3dfy anything in images},
  author={Chen, Xingyu and Chu, Fu-Jen and Gleize, Pierre and Liang, Kevin J and Sax, Alexander and Tang, Hao and Wang, Weiyao and Guo, Michelle and Hardin, Thibaut and Li, Xiang and others},
  journal={arXiv preprint arXiv:2511.16624},
  year={2025}
}

@inproceedings{banerjee2025hot3d,
  title={Hot3d: Hand and object tracking in 3d from egocentric multi-view videos},
  author={Banerjee, Prithviraj and Shkodrani, Sindi and Moulon, Pierre and Hampali, Shreyas and Han, Shangchen and Zhang, Fan and Zhang, Linguang and Fountain, Jade and Miller, Edward and Basol, Selen and others},
  booktitle={Proceedings of the Computer Vision and Pattern Recognition Conference},
  pages={7061--7071},
  year={2025}
}

@inproceedings{christen2024diffh2o,
  title={Diffh2o: Diffusion-based synthesis of hand-object interactions from textual descriptions},
  author={Christen, Sammy and Hampali, Shreyas and Sener, Fadime and Remelli, Edoardo and Hodan, Tomas and Sauser, Eric and Ma, Shugao and Tekin, Bugra},
  booktitle={SIGGRAPH Asia 2024 Conference Papers},
  pages={1--11},
  year={2024}
}

@article{engel2023project,
  title={Project aria: A new tool for egocentric multi-modal ai research},
  author={Engel, Jakob and Somasundaram, Kiran and Goesele, Michael and Sun, Albert and Gamino, Alexander and Turner, Andrew and Talattof, Arjang and Yuan, Arnie and Souti, Bilal and Meredith, Brighid and others},
  journal={arXiv preprint arXiv:2308.13561},
  year={2023}
}

@article{achiam2023gpt,
  title={Gpt-4 technical report},
  author={Achiam, Josh and Adler, Steven and Agarwal, Sandhini and Ahmad, Lama and Akkaya, Ilge and Aleman, Florencia Leoni and Almeida, Diogo and Altenschmidt, Janko and Altman, Sam and Anadkat, Shyamal and others},
  journal={arXiv preprint arXiv:2303.08774},
  year={2023}
}

@article{kerbl20233d,
  title={3D Gaussian splatting for real-time radiance field rendering.},
  author={Kerbl, Bernhard and Kopanas, Georgios and Leimk{\"u}hler, Thomas and Drettakis, George},
  journal={ACM Trans. Graph.},
  volume={42},
  number={4},
  pages={139--1},
  year={2023}
}

@inproceedings{lv2025photoreal,
  title={Photoreal scene reconstruction from an egocentric device},
  author={Lv, Zhaoyang and Monge, Maurizio and Chen, Ka and Zhu, Yufeng and Goesele, Michael and Engel, Jakob and Dong, Zhao and Newcombe, Richard},
  booktitle={Proceedings of the Special Interest Group on Computer Graphics and Interactive Techniques Conference Conference Papers},
  pages={1--11},
  year={2025}
}

@inproceedings{yeshwanth2023scannet++,
  title={Scannet++: A high-fidelity dataset of 3d indoor scenes},
  author={Yeshwanth, Chandan and Liu, Yueh-Cheng and Nie{\ss}ner, Matthias and Dai, Angela},
  booktitle={Proceedings of the IEEE/CVF International Conference on Computer Vision},
  pages={12--22},
  year={2023}
}

@article{ramakrishnan2021habitat,
  title={Habitat-matterport 3d dataset (hm3d): 1000 large-scale 3d environments for embodied ai},
  author={Ramakrishnan, Santhosh K and Gokaslan, Aaron and Wijmans, Erik and Maksymets, Oleksandr and Clegg, Alex and Turner, John and Undersander, Eric and Galuba, Wojciech and Westbury, Andrew and Chang, Angel X and others},
  journal={arXiv preprint arXiv:2109.08238},
  year={2021}
}

@inproceedings{chao2021dexycb,
  title={DexYCB: A benchmark for capturing hand grasping of objects},
  author={Chao, Yu-Wei and Yang, Wei and Xiang, Yu and Molchanov, Pavlo and Handa, Ankur and Tremblay, Jonathan and Narang, Yashraj S and Van Wyk, Karl and Iqbal, Umar and Birchfield, Stan and others},
  booktitle={Proceedings of the IEEE/CVF conference on computer vision and pattern recognition},
  pages={9044--9053},
  year={2021}
}

@article{makoviychuk2021isaac,
  title={Isaac gym: High performance gpu-based physics simulation for robot learning},
  author={Makoviychuk, Viktor and Wawrzyniak, Lukasz and Guo, Yunrong and Lu, Michelle and Storey, Kier and Macklin, Miles and Hoeller, David and Rudin, Nikita and Allshire, Arthur and Handa, Ankur and others},
  journal={arXiv preprint arXiv:2108.10470},
  year={2021}
}

@article{liu2025dextrack,
  title={Dextrack: Towards generalizable neural tracking control for dexterous manipulation from human references},
  author={Liu, Xueyi and Adalibieke, Jianibieke and Han, Qianwei and Qin, Yuzhe and Yi, Li},
  journal={arXiv preprint arXiv:2502.09614},
  year={2025}
}

@article{kim2025pyroki,
  title={PyRoki: A Modular Toolkit for Robot Kinematic Optimization},
  author={Kim, Chung Min and Yi, Brent and Choi, Hongsuk and Ma, Yi and Goldberg, Ken and Kanazawa, Angjoo},
  journal={arXiv preprint arXiv:2505.03728},
  year={2025}
}

@article{li2025scenesplat,
  title={Scenesplat: Gaussian splatting-based scene understanding with vision-language pretraining},
  author={Li, Yue and Ma, Qi and Yang, Runyi and Li, Huapeng and Ma, Mengjiao and Ren, Bin and Popovic, Nikola and Sebe, Nicu and Konukoglu, Ender and Gevers, Theo and others},
  journal={arXiv preprint arXiv:2503.18052},
  year={2025}
}

@inproceedings{li2025latenthoi,
  title={LatentHOI: On the Generalizable Hand Object Motion Generation with Latent Hand Diffusion.},
  author={Li, Muchen and Christen, Sammy and Wan, Chengde and Cai, Yujun and Liao, Renjie and Sigal, Leonid and Ma, Shugao},
  booktitle={Proceedings of the Computer Vision and Pattern Recognition Conference},
  pages={17416--17425},
  year={2025}
}

@inproceedings{petrovich2023tmr,
  title={Tmr: Text-to-motion retrieval using contrastive 3d human motion synthesis},
  author={Petrovich, Mathis and Black, Michael J and Varol, G{\"u}l},
  booktitle={Proceedings of the IEEE/CVF International Conference on Computer Vision},
  pages={9488--9497},
  year={2023}
}

@article{haddadin2024franka,
  title={The franka emika robot: A standard platform in robotics research},
  author={Haddadin, Sami},
  journal={IEEE Robotics \& Automation Magazine},
  year={2024},
  publisher={IEEE}
}

@misc{allegrohand,
  title        = {Allegro Hand v5},
  howpublished = {\url{https://www.allegrohand.com/}},
  note         = {Wonik Robotics},
  year         = {accessed 2026}
}

@inproceedings{lang2025gaussian,
  title={Gaussian-lic: Real-time photo-realistic slam with gaussian splatting and lidar-inertial-camera fusion},
  author={Lang, Xiaolei and Li, Laijian and Wu, Chenming and Zhao, Chen and Liu, Lina and Liu, Yong and Lv, Jiajun and Zuo, Xingxing},
  booktitle={2025 IEEE International Conference on Robotics and Automation (ICRA)},
  pages={8500--8507},
  year={2025},
  organization={IEEE}
}

@misc{metacam,
  title        = {{MetaCam}: {3D} Scanner for Spatial Intelligence},
  author       = {{SkylandX}},
  howpublished = {\url{https://skylandx.com/}},
  year         = {accessed 2026}
}

@article{eldar1997farthest,
  title={The farthest point strategy for progressive image sampling},
  author={Eldar, Yuval and Lindenbaum, Michael and Porat, Moshe and Zeevi, Yehoshua Y},
  journal={IEEE transactions on image processing},
  volume={6},
  number={9},
  pages={1305--1315},
  year={1997},
  publisher={IEEE}
}

@article{li2021learnable,
  title={Learnable fourier features for multi-dimensional spatial positional encoding},
  author={Li, Yang and Si, Si and Li, Gang and Hsieh, Cho-Jui and Bengio, Samy},
  journal={Advances in Neural Information Processing Systems},
  volume={34},
  pages={15816--15829},
  year={2021}
}

@article{hoque2025egodex,
  title={EgoDex: Learning Dexterous Manipulation from Large-Scale Egocentric Video},
  author={Hoque, Ryan and Huang, Peide and Yoon, David J and Sivapurapu, Mouli and Zhang, Jian},
  journal={arXiv preprint arXiv:2505.11709},
  year={2025}
}

@article{MANO:SIGGRAPHASIA:2017,
      title = {Embodied Hands: Modeling and Capturing Hands and Bodies Together},
      author = {Romero, Javier and Tzionas, Dimitrios and Black, Michael J.},
      journal = {ACM Transactions on Graphics, (Proc. SIGGRAPH Asia)},
      volume = {36},
      number = {6},
      series = {245:1--245:17},
      month = nov,
      year = {2017},
      month_numeric = {11}
  }

@article{lipman2022flow,
  title={Flow matching for generative modeling},
  author={Lipman, Yaron and Chen, Ricky TQ and Ben-Hamu, Heli and Nickel, Maximilian and Le, Matt},
  journal={arXiv preprint arXiv:2210.02747},
  year={2022}
}

@article{chen2018neural,
  title={Neural ordinary differential equations},
  author={Chen, Ricky TQ and Rubanova, Yulia and Bettencourt, Jesse and Duvenaud, David K},
  journal={Advances in neural information processing systems},
  volume={31},
  year={2018}
}

@inproceedings{zhou2019continuity,
  title={On the continuity of rotation representations in neural networks},
  author={Zhou, Yi and Barnes, Connelly and Lu, Jingwan and Yang, Jimei and Li, Hao},
  booktitle={Proceedings of the IEEE/CVF conference on computer vision and pattern recognition},
  pages={5745--5753},
  year={2019}
}

@inproceedings{mahmood2019amass,
  title={AMASS: Archive of motion capture as surface shapes},
  author={Mahmood, Naureen and Ghorbani, Nima and Troje, Nikolaus F and Pons-Moll, Gerard and Black, Michael J},
  booktitle={Proceedings of the IEEE/CVF international conference on computer vision},
  pages={5442--5451},
  year={2019}
}

@inproceedings{guo2022generating,
  title={Generating diverse and natural 3d human motions from text},
  author={Guo, Chuan and Zou, Shihao and Zuo, Xinxin and Wang, Sen and Ji, Wei and Li, Xingyu and Cheng, Li},
  booktitle={Proceedings of the IEEE/CVF conference on computer vision and pattern recognition},
  pages={5152--5161},
  year={2022}
}

@inproceedings{potamias2025wilor,
  title={Wilor: End-to-end 3d hand localization and reconstruction in-the-wild},
  author={Potamias, Rolandos Alexandros and Zhang, Jinglei and Deng, Jiankang and Zafeiriou, Stefanos},
  booktitle={Proceedings of the Computer Vision and Pattern Recognition Conference},
  pages={12242--12254},
  year={2025}
}

@inproceedings{radford2021learning,
  title={Learning transferable visual models from natural language supervision},
  author={Radford, Alec and Kim, Jong Wook and Hallacy, Chris and Ramesh, Aditya and Goh, Gabriel and Agarwal, Sandhini and Sastry, Girish and Askell, Amanda and Mishkin, Pamela and Clark, Jack and others},
  booktitle={International conference on machine learning},
  pages={8748--8763},
  year={2021},
  organization={PmLR}
}

@inproceedings{gu2024egolifter,
  title={Egolifter: Open-world 3d segmentation for egocentric perception},
  author={Gu, Qiao and Lv, Zhaoyang and Frost, Duncan and Green, Simon and Straub, Julian and Sweeney, Chris},
  booktitle={European Conference on Computer Vision},
  pages={382--400},
  year={2024},
  organization={Springer}
}

@inproceedings{jiang2024scaling,
  title={Scaling up dynamic human-scene interaction modeling},
  author={Jiang, Nan and Zhang, Zhiyuan and Li, Hongjie and Ma, Xiaoxuan and Wang, Zan and Chen, Yixin and Liu, Tengyu and Zhu, Yixin and Huang, Siyuan},
  booktitle={Proceedings of the IEEE/CVF Conference on Computer Vision and Pattern Recognition},
  pages={1737--1747},
  year={2024}
}

@inproceedings{ghosh2023imos,
  title={IMoS: Intent-Driven Full-Body Motion Synthesis for Human-Object Interactions},
  author={Ghosh, Anindita and Dabral, Rishabh and Golyanik, Vladislav and Theobalt, Christian and Slusallek, Philipp},
  booktitle={Computer Graphics Forum},
  volume={42},
  number={2},
  pages={1--12},
  year={2023},
  organization={Wiley Online Library}
}

@inproceedings{tendulkar2023flex,
  title={Flex: Full-body grasping without full-body grasps},
  author={Tendulkar, Purva and Sur{\'\i}s, D{\'\i}dac and Vondrick, Carl},
  booktitle={Proceedings of the IEEE/CVF Conference on Computer Vision and Pattern Recognition},
  pages={21179--21189},
  year={2023}
}

@inproceedings{liu2022hoi4d,
  title={Hoi4d: A 4d egocentric dataset for category-level human-object interaction},
  author={Liu, Yunze and Liu, Yun and Jiang, Che and Lyu, Kangbo and Wan, Weikang and Shen, Hao and Liang, Boqiang and Fu, Zhoujie and Wang, He and Yi, Li},
  booktitle={Proceedings of the IEEE/CVF Conference on Computer Vision and Pattern Recognition},
  pages={21013--21022},
  year={2022}
}

@inproceedings{fan2023arctic,
  title={ARCTIC: A dataset for dexterous bimanual hand-object manipulation},
  author={Fan, Zicong and Taheri, Omid and Tzionas, Dimitrios and Kocabas, Muhammed and Kaufmann, Manuel and Black, Michael J and Hilliges, Otmar},
  booktitle={Proceedings of the IEEE/CVF conference on computer vision and pattern recognition},
  pages={12943--12954},
  year={2023}
}

@inproceedings{li2024task,
  title={Task-oriented human-object interactions generation with implicit neural representations},
  author={Li, Quanzhou and Wang, Jingbo and Loy, Chen Change and Dai, Bo},
  booktitle={Proceedings of the IEEE/CVF Winter Conference on Applications of Computer Vision},
  pages={3035--3044},
  year={2024}
}

@inproceedings{hassan2023synthesizing,
  title={Synthesizing physical character-scene interactions},
  author={Hassan, Mohamed and Guo, Yunrong and Wang, Tingwu and Black, Michael and Fidler, Sanja and Peng, Xue Bin},
  booktitle={ACM SIGGRAPH 2023 Conference Proceedings},
  pages={1--9},
  year={2023}
}

@inproceedings{grady2021contactopt,
  title={Contactopt: Optimizing contact to improve grasps},
  author={Grady, Patrick and Tang, Chengcheng and Twigg, Christopher D and Vo, Minh and Brahmbhatt, Samarth and Kemp, Charles C},
  booktitle={Proceedings of the IEEE/CVF Conference on Computer Vision and Pattern Recognition},
  pages={1471--1481},
  year={2021}
}

@inproceedings{karunratanakul2020grasping,
  title={Grasping field: Learning implicit representations for human grasps},
  author={Karunratanakul, Korrawe and Yang, Jinlong and Zhang, Yan and Black, Michael J and Muandet, Krikamol and Tang, Siyu},
  booktitle={2020 International Conference on 3D Vision (3DV)},
  pages={333--344},
  year={2020},
  organization={IEEE}
}

@inproceedings{jiang2021hand,
  title={Hand-object contact consistency reasoning for human grasps generation},
  author={Jiang, Hanwen and Liu, Shaowei and Wang, Jiashun and Wang, Xiaolong},
  booktitle={Proceedings of the IEEE/CVF international conference on computer vision},
  pages={11107--11116},
  year={2021}
}

@inproceedings{taheri2022goal,
  title={GOAL: Generating 4D whole-body motion for hand-object grasping},
  author={Taheri, Omid and Choutas, Vasileios and Black, Michael J and Tzionas, Dimitrios},
  booktitle={Proceedings of the IEEE/CVF Conference on Computer Vision and Pattern Recognition},
  pages={13263--13273},
  year={2022}
}

@article{zhang2021manipnet,
  title={Manipnet: neural manipulation synthesis with a hand-object spatial representation},
  author={Zhang, He and Ye, Yuting and Shiratori, Takaaki and Komura, Taku},
  journal={ACM Transactions on Graphics (ToG)},
  volume={40},
  number={4},
  pages={1--14},
  year={2021},
  publisher={ACM New York, NY, USA}
}

@inproceedings{zheng2023cams,
  title={Cams: Canonicalized manipulation spaces for category-level functional hand-object manipulation synthesis},
  author={Zheng, Juntian and Zheng, Qingyuan and Fang, Lixing and Liu, Yun and Yi, Li},
  booktitle={Proceedings of the IEEE/CVF Conference on Computer Vision and Pattern Recognition},
  pages={585--594},
  year={2023}
}

@inproceedings{braun2024physically,
  title={Physically plausible full-body hand-object interaction synthesis},
  author={Braun, Jona and Christen, Sammy and Kocabas, Muhammed and Aksan, Emre and Hilliges, Otmar},
  booktitle={2024 International Conference on 3D Vision (3DV)},
  pages={464--473},
  year={2024},
  organization={IEEE}
}

@article{rajeswaran2017learning,
  title={Learning complex dexterous manipulation with deep reinforcement learning and demonstrations},
  author={Rajeswaran, Aravind and Kumar, Vikash and Gupta, Abhishek and Vezzani, Giulia and Schulman, John and Todorov, Emanuel and Levine, Sergey},
  journal={arXiv preprint arXiv:1709.10087},
  year={2017}
}

@article{liu2024geneoh,
  title={Geneoh diffusion: Towards generalizable hand-object interaction denoising via denoising diffusion},
  author={Liu, Xueyi and Yi, Li},
  journal={arXiv preprint arXiv:2402.14810},
  year={2024}
}

@inproceedings{cha2024text2hoi,
  title={Text2hoi: Text-guided 3d motion generation for hand-object interaction},
  author={Cha, Junuk and Kim, Jihyeon and Yoon, Jae Shin and Baek, Seungryul},
  booktitle={Proceedings of the IEEE/CVF Conference on Computer Vision and Pattern Recognition},
  pages={1577--1585},
  year={2024}
}

@article{tevet2022human,
  title={Human motion diffusion model},
  author={Tevet, Guy and Raab, Sigal and Gordon, Brian and Shafir, Yonatan and Cohen-Or, Daniel and Bermano, Amit H},
  journal={arXiv preprint arXiv:2209.14916},
  year={2022}
}

@inproceedings{petrovich2022temos,
  title={Temos: Generating diverse human motions from textual descriptions},
  author={Petrovich, Mathis and Black, Michael J and Varol, G{\"u}l},
  booktitle={European Conference on Computer Vision},
  pages={480--497},
  year={2022},
  organization={Springer}
}

@inproceedings{chen2023executing,
  title={Executing your commands via motion diffusion in latent space},
  author={Chen, Xin and Jiang, Biao and Liu, Wen and Huang, Zilong and Fu, Bin and Chen, Tao and Yu, Gang},
  booktitle={Proceedings of the IEEE/CVF conference on computer vision and pattern recognition},
  pages={18000--18010},
  year={2023}
}

@inproceedings{karunratanakul2023guided,
  title={Guided motion diffusion for controllable human motion synthesis},
  author={Karunratanakul, Korrawe and Preechakul, Konpat and Suwajanakorn, Supasorn and Tang, Siyu},
  booktitle={Proceedings of the IEEE/CVF International Conference on Computer Vision},
  pages={2151--2162},
  year={2023}
}

@article{zhang2024motiondiffuse,
  title={Motiondiffuse: Text-driven human motion generation with diffusion model},
  author={Zhang, Mingyuan and Cai, Zhongang and Pan, Liang and Hong, Fangzhou and Guo, Xinying and Yang, Lei and Liu, Ziwei},
  journal={IEEE transactions on pattern analysis and machine intelligence},
  volume={46},
  number={6},
  pages={4115--4128},
  year={2024},
  publisher={IEEE}
}

@inproceedings{kwon2021h2o,
  title={H2o: Two hands manipulating objects for first person interaction recognition},
  author={Kwon, Taein and Tekin, Bugra and St{\"u}hmer, Jan and Bogo, Federica and Pollefeys, Marc},
  booktitle={Proceedings of the IEEE/CVF international conference on computer vision},
  pages={10138--10148},
  year={2021}
}

@article{li2023object,
  title={Object motion guided human motion synthesis},
  author={Li, Jiaman and Wu, Jiajun and Liu, C Karen},
  journal={ACM Transactions on Graphics (TOG)},
  volume={42},
  number={6},
  pages={1--11},
  year={2023},
  publisher={ACM New York, NY, USA}
}

@inproceedings{bhatnagar22behave,
title = {BEHAVE: Dataset and Method for Tracking Human Object Interactions},
author={Bhatnagar, Bharat Lal and Xie, Xianghui and Petrov, Ilya and Sminchisescu, Cristian and Theobalt, Christian and Pons-Moll, Gerard},
booktitle = {{IEEE} Conference on Computer Vision and Pattern Recognition (CVPR)},
month = {jun},
organization = {{IEEE}},
year = {2022},
}

@article{lu2025humoto,
  title={HUMOTO: A 4D Dataset of Mocap Human Object Interactions},
  author={Lu, Jiaxin and Huang, Chun-Hao Paul and Bhattacharya, Uttaran and Huang, Qixing and Zhou, Yi},
  journal={arXiv preprint arXiv:2504.10414},
  year={2025}
}

@inproceedings{xu2023interdiff,
  title={Interdiff: Generating 3d human-object interactions with physics-informed diffusion},
  author={Xu, Sirui and Li, Zhengyuan and Wang, Yu-Xiong and Gui, Liang-Yan},
  booktitle={Proceedings of the IEEE/CVF International Conference on Computer Vision},
  pages={14928--14940},
  year={2023}
}

@inproceedings{li2024controllable,
  title={Controllable human-object interaction synthesis},
  author={Li, Jiaman and Clegg, Alexander and Mottaghi, Roozbeh and Wu, Jiajun and Puig, Xavier and Liu, C Karen},
  booktitle={European Conference on Computer Vision},
  pages={54--72},
  year={2024},
  organization={Springer}
}

@inproceedings{peng2025hoi,
  title={Hoi-diff: Text-driven synthesis of 3d human-object interactions using diffusion models},
  author={Peng, Xiaogang and Xie, Yiming and Wu, Zizhao and Jampani, Varun and Sun, Deqing and Jiang, Huaizu},
  booktitle={Proceedings of the Computer Vision and Pattern Recognition Conference},
  pages={2878--2888},
  year={2025}
}

@article{wang2023physhoi,
  title={Physhoi: Physics-based imitation of dynamic human-object interaction},
  author={Wang, Yinhuai and Lin, Jing and Zeng, Ailing and Luo, Zhengyi and Zhang, Jian and Zhang, Lei},
  journal={arXiv preprint arXiv:2312.04393},
  year={2023}
}

@inproceedings{mandlekar2019scaling,
  title={Scaling robot supervision to hundreds of hours with roboturk: Robotic manipulation dataset through human reasoning and dexterity},
  author={Mandlekar, Ajay and Booher, Jonathan and Spero, Max and Tung, Albert and Gupta, Anchit and Zhu, Yuke and Garg, Animesh and Savarese, Silvio and Fei-Fei, Li},
  booktitle={2019 IEEE/RSJ International Conference on Intelligent Robots and Systems (IROS)},
  pages={1048--1055},
  year={2019},
  organization={IEEE}
}

@article{khazatsky2024droid,
  title={Droid: A large-scale in-the-wild robot manipulation dataset},
  author={Khazatsky, Alexander and Pertsch, Karl and Nair, Suraj and Balakrishna, Ashwin and Dasari, Sudeep and Karamcheti, Siddharth and Nasiriany, Soroush and Srirama, Mohan Kumar and Chen, Lawrence Yunliang and Ellis, Kirsty and others},
  journal={arXiv preprint arXiv:2403.12945},
  year={2024}
}

@inproceedings{o2024open,
  title={Open x-embodiment: Robotic learning datasets and rt-x models: Open x-embodiment collaboration 0},
  author={O’Neill, Abby and Rehman, Abdul and Maddukuri, Abhiram and Gupta, Abhishek and Padalkar, Abhishek and Lee, Abraham and Pooley, Acorn and Gupta, Agrim and Mandlekar, Ajay and Jain, Ajinkya and others},
  booktitle={2024 IEEE International Conference on Robotics and Automation (ICRA)},
  pages={6892--6903},
  year={2024},
  organization={IEEE}
}

@inproceedings{perrett2025hd,
  title={Hd-epic: A highly-detailed egocentric video dataset},
  author={Perrett, Toby and Darkhalil, Ahmad and Sinha, Saptarshi and Emara, Omar and Pollard, Sam and Parida, Kranti Kumar and Liu, Kaiting and Gatti, Prajwal and Bansal, Siddhant and Flanagan, Kevin and others},
  booktitle={Proceedings of the Computer Vision and Pattern Recognition Conference},
  pages={23901--23913},
  year={2025}
}

@article{damen2020epic,
  title={The epic-kitchens dataset: Collection, challenges and baselines},
  author={Damen, Dima and Doughty, Hazel and Farinella, Giovanni Maria and Fidler, Sanja and Furnari, Antonino and Kazakos, Evangelos and Moltisanti, Davide and Munro, Jonathan and Perrett, Toby and Price, Will and others},
  journal={IEEE Transactions on Pattern Analysis and Machine Intelligence},
  volume={43},
  number={11},
  pages={4125--4141},
  year={2020},
  publisher={IEEE}
}

@article{kim2024openvla,
  title={Openvla: An open-source vision-language-action model},
  author={Kim, Moo Jin and Pertsch, Karl and Karamcheti, Siddharth and Xiao, Ted and Balakrishna, Ashwin and Nair, Suraj and Rafailov, Rafael and Foster, Ethan and Lam, Grace and Sanketi, Pannag and others},
  journal={arXiv preprint arXiv:2406.09246},
  year={2024}
}

@misc{black2026pi0visionlanguageactionflowmodel,
      title={$\pi_0$: A Vision-Language-Action Flow Model for General Robot Control}, 
      author={Kevin Black and Noah Brown and Danny Driess and Adnan Esmail and Michael Equi and Chelsea Finn and Niccolo Fusai and Lachy Groom and Karol Hausman and Brian Ichter and Szymon Jakubczak and Tim Jones and Liyiming Ke and Sergey Levine and Adrian Li-Bell and Mohith Mothukuri and Suraj Nair and Karl Pertsch and Lucy Xiaoyang Shi and James Tanner and Quan Vuong and Anna Walling and Haohuan Wang and Ury Zhilinsky},
      year={2026},
      eprint={2410.24164},
      archivePrefix={arXiv},
      primaryClass={cs.LG},
      url={https://arxiv.org/abs/2410.24164}, 
}

@article{wang2025unified,
  title={Unified Vision-Language-Action Model},
  author={Wang, Yuqi and Li, Xinghang and Wang, Wenxuan and Zhang, Junbo and Li, Yingyan and Chen, Yuntao and Wang, Xinlong and Zhang, Zhaoxiang},
  journal={arXiv preprint arXiv:2506.19850},
  year={2025}
}

@article{ahn2022can,
  title={Do as i can, not as i say: Grounding language in robotic affordances},
  author={Ahn, Michael and Brohan, Anthony and Brown, Noah and Chebotar, Yevgen and Cortes, Omar and David, Byron and Finn, Chelsea and Fu, Chuyuan and Gopalakrishnan, Keerthana and Hausman, Karol and others},
  journal={arXiv preprint arXiv:2204.01691},
  year={2022}
}

@article{hao2024language,
  title={Language-guided manipulation with diffusion policies and constrained inpainting},
  author={Hao, Ce and Lin, Kelvin and Luo, Siyuan and Soh, Harold},
  journal={arXiv preprint arXiv:2406.09767},
  year={2024}
}

@article{goswami2025world,
  title={World Models for Learning Dexterous Hand-Object Interactions from Human Videos},
  author={Goswami, Raktim Gautam and Bar, Amir and Fan, David and Yang, Tsung-Yen and Zhou, Gaoyue and Krishnamurthy, Prashanth and Rabbat, Michael and Khorrami, Farshad and LeCun, Yann},
  journal={arXiv preprint arXiv:2512.13644},
  year={2025}
}

@article{bharadhwaj2024gen2act,
  title={Gen2act: Human video generation in novel scenarios enables generalizable robot manipulation},
  author={Bharadhwaj, Homanga and Dwibedi, Debidatta and Gupta, Abhinav and Tulsiani, Shubham and Doersch, Carl and Xiao, Ted and Shah, Dhruv and Xia, Fei and Sadigh, Dorsa and Kirmani, Sean},
  journal={arXiv preprint arXiv:2409.16283},
  year={2024}
}

@article{mao2025robot,
  title={Robot Learning from a Physical World Model},
  author={Mao, Jiageng and He, Sicheng and Wu, Hao-Ning and You, Yang and Sun, Shuyang and Wang, Zhicheng and Bao, Yanan and Chen, Huizhong and Guibas, Leonidas and Guizilini, Vitor and others},
  journal={arXiv preprint arXiv:2511.07416},
  year={2025}
}

@inproceedings{wang2025language,
  title={This\&that: Language-gesture controlled video generation for robot planning},
  author={Wang, Boyang and Sridhar, Nikhil and Feng, Chao and Van der Merwe, Mark and Fishman, Adam and Fazeli, Nima and Park, Jeong Joon},
  booktitle={2025 IEEE International Conference on Robotics and Automation (ICRA)},
  pages={12842--12849},
  year={2025},
  organization={IEEE}
}

@article{alayrac2022flamingo,
  title={Flamingo: a visual language model for few-shot learning},
  author={Alayrac, Jean-Baptiste and Donahue, Jeff and Luc, Pauline and Miech, Antoine and Barr, Iain and Hasson, Yana and Lenc, Karel and Mensch, Arthur and Millican, Katherine and Reynolds, Malcolm and others},
  journal={Advances in neural information processing systems},
  volume={35},
  pages={23716--23736},
  year={2022}
}

@article{peng2023kosmos,
  title={Kosmos-2: Grounding multimodal large language models to the world},
  author={Peng, Zhiliang and Wang, Wenhui and Dong, Li and Hao, Yaru and Huang, Shaohan and Ma, Shuming and Wei, Furu},
  journal={arXiv preprint arXiv:2306.14824},
  year={2023}
}

@article{beyer2024paligemma,
  title={Paligemma: A versatile 3b vlm for transfer},
  author={Beyer, Lucas and Steiner, Andreas and Pinto, Andr{\'e} Susano and Kolesnikov, Alexander and Wang, Xiao and Salz, Daniel and Neumann, Maxim and Alabdulmohsin, Ibrahim and Tschannen, Michael and Bugliarello, Emanuele and others},
  journal={arXiv preprint arXiv:2407.07726},
  year={2024}
}

@article{li2025eagle,
  title={Eagle 2: Building post-training data strategies from scratch for frontier vision-language models},
  author={Li, Zhiqi and Chen, Guo and Liu, Shilong and Wang, Shihao and VS, Vibashan and Ji, Yishen and Lan, Shiyi and Zhang, Hao and Zhao, Yilin and Radhakrishnan, Subhashree and others},
  journal={arXiv preprint arXiv:2501.14818},
  year={2025}
}

@book{Gibson1979-GIBTEA,
	author = {James J. Gibson},
	editor = {},
	publisher = {Houghton Mifflin},
	title = {The Ecological Approach to Visual Perception: Classic Edition},
	year = {1979}
}

@book{mason2001mechanics,
  title={Mechanics of robotic manipulation},
  author={Mason, Matthew T},
  year={2001},
  publisher={MIT press}
}

@inproceedings{bicchi2000robotic,
  title={Robotic grasping and contact: A review},
  author={Bicchi, Antonio and Kumar, Vijay},
  booktitle={Proceedings 2000 ICRA. Millennium conference. IEEE international conference on robotics and automation. Symposia proceedings (Cat. No. 00CH37065)},
  volume={1},
  pages={348--353},
  year={2000},
  organization={IEEE}
}

@inproceedings{pan2025omnimanip,
  title={Omnimanip: Towards general robotic manipulation via object-centric interaction primitives as spatial constraints},
  author={Pan, Mingjie and Zhang, Jiyao and Wu, Tianshu and Zhao, Yinghao and Gao, Wenlong and Dong, Hao},
  booktitle={Proceedings of the Computer Vision and Pattern Recognition Conference},
  pages={17359--17369},
  year={2025}
}

@article{xu2025a0,
  title={A0: An affordance-aware hierarchical model for general robotic manipulation},
  author={Xu, Rongtao and Zhang, Jian and Guo, Minghao and Wen, Youpeng and Yang, Haoting and Lin, Min and Huang, Jianzheng and Li, Zhe and Zhang, Kaidong and Wang, Liqiong and others},
  journal={arXiv preprint arXiv:2504.12636},
  year={2025}
}

@inproceedings{hsu2025spot,
  title={Spot: Se (3) pose trajectory diffusion for object-centric manipulation},
  author={Hsu, Cheng-Chun and Wen, Bowen and Xu, Jie and Narang, Yashraj and Wang, Xiaolong and Zhu, Yuke and Biswas, Joydeep and Birchfield, Stan},
  booktitle={2025 IEEE International Conference on Robotics and Automation (ICRA)},
  pages={4853--4860},
  year={2025},
  organization={IEEE}
}

@article{lum2025crossing,
  title={Crossing the human-robot embodiment gap with sim-to-real rl using one human demonstration},
  author={Lum, Tyler Ga Wei and Lee, Olivia Y and Liu, C Karen and Bohg, Jeannette},
  journal={arXiv preprint arXiv:2504.12609},
  year={2025}
}

@article{fu2026egograsp,
  title={EgoGrasp: World-Space Hand-Object Interaction Estimation from Egocentric Videos},
  author={Fu, Hongming and Wang, Wenjia and Qiao, Xiaozhen and Potamias, Rolandos Alexandros and Komura, Taku and Yang, Shuo and Liu, Zheng and Zhao, Bo},
  journal={arXiv preprint arXiv:2601.01050},
  year={2026}
}

@inproceedings{xu2023egopca,
  title={Egopca: A new framework for egocentric hand-object interaction understanding},
  author={Xu, Yue and Li, Yong-Lu and Huang, Zhemin and Liu, Michael Xu and Lu, Cewu and Tai, Yu-Wing and Tang, Chi-Keung},
  booktitle={Proceedings of the IEEE/CVF International Conference on Computer Vision},
  pages={5273--5284},
  year={2023}
}

@article{song2020denoising,
  title={Denoising diffusion implicit models},
  author={Song, Jiaming and Meng, Chenlin and Ermon, Stefano},
  journal={arXiv preprint arXiv:2010.02502},
  year={2020}
}

@article{ho2020denoising,
  title={Denoising diffusion probabilistic models},
  author={Ho, Jonathan and Jain, Ajay and Abbeel, Pieter},
  journal={Advances in neural information processing systems},
  volume={33},
  pages={6840--6851},
  year={2020}
}

@article{johansson2009coding,
  title={Coding and use of tactile signals from the fingertips in object manipulation tasks},
  author={Johansson, Roland S and Flanagan, J Randall},
  journal={Nature Reviews Neuroscience},
  volume={10},
  number={5},
  pages={345--359},
  year={2009},
  publisher={Nature Publishing Group UK London}
}

@article{oord2018representation,
  title={Representation learning with contrastive predictive coding},
  author={Oord, Aaron van den and Li, Yazhe and Vinyals, Oriol},
  journal={arXiv preprint arXiv:1807.03748},
  year={2018}
}

@article{qi2017pointnet++,
  title={Pointnet++: Deep hierarchical feature learning on point sets in a metric space},
  author={Qi, Charles Ruizhongtai and Yi, Li and Su, Hao and Guibas, Leonidas J},
  journal={Advances in neural information processing systems},
  volume={30},
  year={2017}
}

@inproceedings{hu2018squeeze,
  title={Squeeze-and-excitation networks},
  author={Hu, Jie and Shen, Li and Sun, Gang},
  booktitle={Proceedings of the IEEE conference on computer vision and pattern recognition},
  pages={7132--7141},
  year={2018}
}

@article{paszke2019pytorch,
  title={Pytorch: An imperative style, high-performance deep learning library},
  author={Paszke, Adam and Gross, Sam and Massa, Francisco and Lerer, Adam and Bradbury, James and Chanan, Gregory and Killeen, Trevor and Lin, Zeming and Gimelshein, Natalia and Antiga, Luca and others},
  journal={Advances in neural information processing systems},
  volume={32},
  year={2019}
}

@article{zhong2025dexgraspvla,
  title={Dexgraspvla: A vision-language-action framework towards general dexterous grasping},
  author={Zhong, Yifan and Huang, Xuchuan and Li, Ruochong and Zhang, Ceyao and Chen, Zhang and Guan, Tianrui and Zeng, Fanlian and Lui, Ka Num and Ye, Yuyao and Liang, Yitao and others},
  journal={arXiv preprint arXiv:2502.20900},
  year={2025}
}

@article{ye2026whole,
  title={Whole: World-grounded hand-object lifted from egocentric videos},
  author={Ye, Yufei and Li, Jiaman and Rong, Ryan and Liu, C Karen},
  journal={arXiv preprint arXiv:2602.22209},
  year={2026}
}

@inproceedings{petrov2025tridi,
  title={Tridi: Trilateral diffusion of 3d humans, objects, and interactions},
  author={Petrov, Ilya A and Marin, Riccardo and Chibane, Julian and Pons-Moll, Gerard},
  booktitle={Proceedings of the IEEE/CVF International Conference on Computer Vision},
  pages={5523--5535},
  year={2025}
}

@article{saroha2026egoflow,
  title={EgoFlow: Gradient-Guided Flow Matching for Egocentric 6DoF Object Motion Generation},
  author={Saroha, Abhishek and Zeng, Huajian and Zuo, Xingxing and Cremers, Daniel and Wang, Xi},
  journal={arXiv preprint arXiv:2604.01421},
  year={2026}
}

@inproceedings{zeng2026gmt,
  title={GMT: Goal-conditioned multimodal transformer for 6-dof object trajectory synthesis in 3d scenes},
  author={Zeng, Huajian and Saroha, Abhishek and Cremers, Daniel and Wang, Xi},
  booktitle={2026 International Conference on 3D Vision (3DV)},
  pages={1691--1701},
  year={2026},
  organization={IEEE}
}

@article{zeng2026stablehand,
  title={StableHand: Quality-Aware Flow Matching for World-Space Dual-Hand Motion Estimation from Egocentric Video},
  author={Zeng, Huajian and Yao, Chaohua and Zhang, Yuantai and Yang, Jiaqi and Potamias, Rolandos Alexandros and Zuo, Xingxing},
  journal={arXiv preprint arXiv:2605.18553},
  year={2026}
}

@inproceedings{zhang2025hawor,
  title={Hawor: World-space hand motion reconstruction from egocentric videos},
  author={Zhang, Jinglei and Deng, Jiankang and Ma, Chao and Potamias, Rolandos Alexandros},
  booktitle={Proceedings of the Computer Vision and Pattern Recognition Conference},
  pages={1805--1815},
  year={2025}
}

\clearpage
\appendix


\begin{center}
  {\LARGE\bfseries Supplementary Material}
\end{center}
\bigskip


This supplementary document provides additional details and results that complement the main paper. We begin with implementation details of the model architecture and training configurations in Sec.~\ref{sec:supp_implementation}, followed by an ablation study on the flow matching prediction target in Sec.~\ref{sec:supp_prediction_target}. We then describe the hand-object data reconstruction pipeline\rev{, together with the effect of EgoDex pretraining,} and dataset preprocessing in Sec.~\ref{sec:supp_data_reconstruction} and Sec.~\ref{sec:supp_dataset}, respectively. Next, we elaborate on the physics simulation evaluation protocol in Sec.~\ref{sec:supp_simulation} and provide details on the real-world experiment setup in Sec.~\ref{sec:supp_realworld}. We then analyze the scene representation\rev{, the granularity of scene conditioning, and hand-scene penetration under clutter} in Sec.~\ref{sec:supp_scene}. Full evaluation metric definitions are provided in Sec.~\ref{sec:supp_metrics}. We additionally provide a stage-transition soft-inpainting ablation in Sec.~\ref{sec:supp_inpainting}, \rev{a comparison of single-stage and two-stage generation in Sec.~\ref{sec:supp_single_stage}, a post-grasp dexterity analysis in Sec.~\ref{sec:supp_dexterity},} additional qualitative results in Sec.~\ref{sec:supp_qualitative}, and failure cases in Sec.~\ref{sec:supp_failure}. We also provide a supplementary video with animated results.

\section{Implementation details}
\label{sec:supp_implementation}

This section provides additional implementation details beyond Sec.~\ref{subsec:implementation_details} of the main paper.

\subsection{Flow matching: deferred definitions}
\label{subsec:supp_fm_details}

We collect here the flow matching definitions deferred from Sec.~\ref{sec:preliminary} of the main paper.

\noindent\textbf{Neural ODE formulation.}
The continuous-time transport from $p_0$ to $p_1$ underlying conditional flow matching is defined by the neural ODE~\cite{chen2018neural}:
\begin{equation}
\frac{d}{d\tau}\,\boldsymbol{\phi}_\tau(\mathbf{x}) = \mathbf{v}\!\left(\boldsymbol{\phi}_\tau(\mathbf{x}),\tau\right),
\qquad \boldsymbol{\phi}_0(\mathbf{x})=\mathbf{x},
\label{eq:fm_ode}
\end{equation}
where $\boldsymbol{\phi}_\tau$ is the flow map. The optimal-transport path of Eq.~\ref{eq:cfm_path} is one specific instantiation of this ODE.

\noindent\textbf{Role of $\sigma_{\min}$.}
The constant $\sigma_{\min}>0$ in Eq.~\ref{eq:cfm_path} controls the residual stochasticity at the end of the flow: at $\tau=1$ the conditional distribution becomes $p_1(\mathbf{x}\mid\mathbf{x}_1)=\mathcal{N}(\mathbf{x}_1,\sigma_{\min}^2\mathbf{I})$ rather than a Dirac delta, preventing the path from degenerating and keeping the closed-form vector field of Eq.~\ref{eq:cfm_target} numerically stable as $\tau\to 1$. We set $\sigma_{\min}=10^{-5}$, small enough that its effect on sample quality is negligible.

\noindent\textbf{Euler discretization at inference.}
At inference, sampling starts from $\mathbf{x}(0)=\mathbf{x}_0\sim p_0$ and integrates Eq.~\ref{eq:fm_ode} to $\tau=1$ with a numerical ODE solver. We use Euler discretization with $K=50$ steps:
\begin{equation}
\mathbf{x}_{k+1} = \mathbf{x}_k + \Delta \tau\ \mathbf{v}_\theta(\mathbf{x}_k,\tau_k,\mathbf{c}),
\qquad \Delta \tau=\tfrac{1}{K},\ \ \tau_k=\tfrac{k}{K}.
\label{eq:euler}
\end{equation}

\subsection{Architecture and training}
\label{subsec:supp_architecture}

Both models use a DiT~\cite{peebles2023scalable} backbone with pre-norm.
The manipulation model uses $d{=}512$, $d_{\text{ff}}{=}1024$, while the grasping model uses $d{=}256$, $d_{\text{ff}}{=}512$. Both use $L{=}8$ layers and dropout $0.1$.
Text is encoded by a frozen T5-Large~\cite{raffel2020exploring}, projected to $d$ and injected via cross-attention.
Object geometry uses BPS~\cite{prokudin2019efficient} encoding, projected to 256-dim and concatenated per-frame.

The scene-tokenization pipeline below is consumed only by the manipulation stage; the grasping stage is geometry-centric (cf.\ Sec.~\ref{subsec:two_stage_generation}) and does not include scene tokens in its conditioning. The scene is represented by up to 25{,}000 points sampled via Farthest Point Sampling (FPS)~\cite{eldar1997farthest}. Geometric features from Concerto~\cite{zhang2025concerto} $d_e{=}1536$ and semantic features from SceneSplat~\cite{li2025scenesplat} $d_u{=}768$ are fused via bidirectional cross-attention with $d_h{=}512$.
3D coordinates use Fourier positional encoding~\cite{li2021learnable}, projected to 64-dim.
Fused features are compressed by a Perceiver~\cite{jaegle2021perceiver} bottleneck into $K{=}256$ scene tokens of dimension $d$.
For global scene encoding, we voxelize the scene at $48{\times}48{\times}24$ resolution and process it with a ViT, injected via AdaLN~\cite{peebles2023scalable}.

We use AdamW~\cite{loshchilov2017decoupled} with lr $10^{-4}$, weight decay $0.01$, and linear annealing.
The grasping model trains for 500K steps, and the manipulation model for 200K steps.
We adopt logit-normal time sampling and classifier-free guidance with drop probability $0.1$ during training and scale $2.5$ at inference with 50-step Euler integration.

Each frame is a 117-dim vector encoding bimanual wrist poses, hand articulations in 24D PCA, per-joint signed distances to the object, and object pose.
The full sequence length is $N{=}200$ frames with $N_g{=}50$ for grasping and $150$ for manipulation.

\noindent\textbf{Contrastive Alignment Loss.}
We employ a TMR-inspired~\cite{petrovich2023tmr} symmetric InfoNCE loss to align motion and text embeddings in a shared 512-dim latent space. The motion encoder is a 4-layer Transformer encoder with a learnable \texttt{[CLS]} token. The text encoder applies mean pooling over frozen T5-Large~\cite{raffel2020exploring} embeddings and projects them to the same 512-dim space via a two-layer MLP. The temperature $\tau$ is initialized to $0.7$, learnable, and clamped to $[0.01, 100]$. The alignment loss is weighted by $\lambda_{\text{align}}{=}0.1$ and directly added to the flow matching loss. These alignment components are only used during training and discarded at inference.

\noindent\textbf{Baseline reproduction.}
For DiffH2O~\cite{christen2024diffh2o} and LatentHOI~\cite{li2025latenthoi} we use the official codebases and follow each paper's published training schedule (optimizer, learning rate, batch size, total iterations) on GRAB and HOT3D, without re-tuning. The only deviation is the CLIP~\cite{radford2021learning} $\to$ T5-Large~\cite{raffel2020exploring} text encoder swap applied uniformly for fairness; we re-initialize the linear projection that maps text features into each model's hidden dimension. All other hyperparameters (\eg, diffusion steps and CFG scale for DiffH2O, latent dimensionality and decoder depth for LatentHOI) follow the original papers.


\section{Flow matching prediction target analysis}
\label{sec:supp_prediction_target}

Different from standard flow matching that predicts the velocity field $\mathbf{v}$,
we propose to directly predict the clean data $\mathbf{x}_1$ as the model output.
Recent studies~\cite{li2025back} have conducted extensive analyses on the choice of prediction targets and consistently advocate for directly predicting the clean data ($x$-prediction), rather than noise ($\epsilon$-prediction) or flow velocity ($v$-prediction).
These works are motivated by the \emph{manifold assumption}, which posits that natural data, such as images, concentrate on a low-dimensional manifold embedded in a high-dimensional ambient space, while quantities such as noise $\epsilon$ or flow velocity $v = x - \epsilon$ are inherently off-manifold and distributed across the full ambient space.

\begin{figure}[t]
    \centering
    \includegraphics[width=\linewidth]{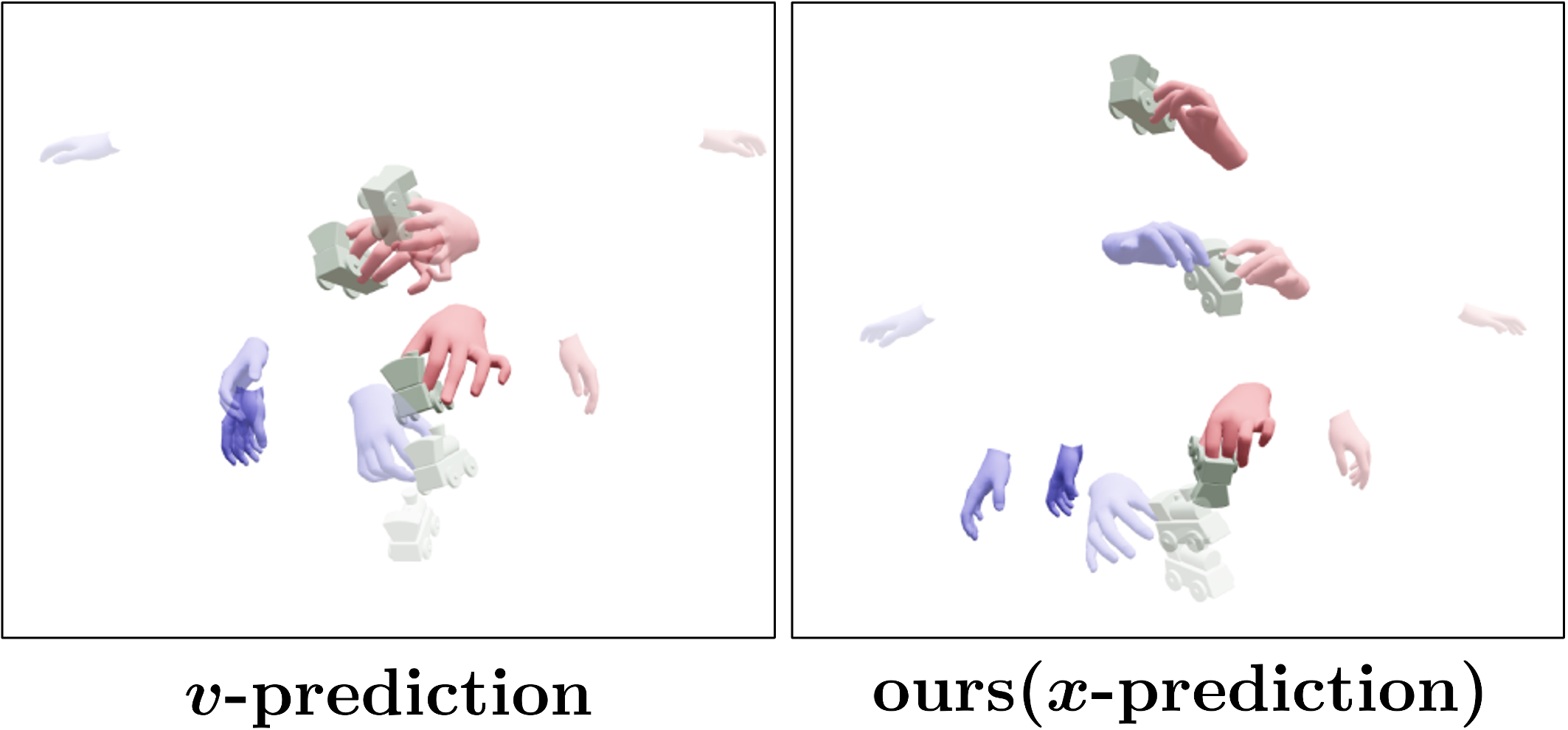}
    \caption{Qualitative comparison of prediction targets. Our $x$-prediction produces temporally smooth and stable hand trajectories, while $v$-prediction exhibits noticeable spatial jitter and inconsistent grasping poses across frames.}
    \label{fig:prediction_target_vis}
\end{figure}

In our problem, although the interaction sequence is represented in a high-dimensional space, physically and semantically plausible hand-object motions are highly constrained (e.g., kinematics, joint limits, contact consistency, and temporal coherence), suggesting a similarly thin \emph{interaction manifold}.
Under velocity-based training, small high-frequency errors in $\mathbf{v}_\theta$ are directly injected into the ODE integration and can accumulate across steps, manifesting as temporal jitter~\cite{li2025back}. In contrast, $x$-prediction repeatedly estimates an on-manifold state and then derives the vector field from it, which empirically improves temporal stability.
To validate this hypothesis, we conduct an ablation study comparing the two prediction targets ($x$-prediction and $v$-prediction) on the GRAB~\cite{taheri2020grab} dataset. We evaluate temporal smoothness using two complementary metrics (see Sec.~\ref{sec:supp_metrics} for formal definitions): (1)~\emph{jerk}, the third derivative of object position/rotation with respect to time, and (2)~\emph{hand acceleration}, which measures the second-order temporal stability of the generated hand motion, decomposed into global wrist acceleration ($\mathrm{Acc}_g$) and local finger acceleration ($\mathrm{Acc}_l$) for both positional and rotational components.
As shown in Tab.~\ref{tab:ablation_prediction_target}, $x$-prediction achieves lower angular jerk and consistently lower local hand acceleration (both positional and rotational), indicating smoother finger articulation. The global wrist metrics are comparable between the two targets, suggesting that the primary benefit of $x$-prediction lies in stabilizing fine-grained hand dynamics rather than coarse wrist trajectories.
Fig.~\ref{fig:prediction_target_vis} further visualizes the generated trajectories, where $v$-prediction exhibits noticeable high-frequency jitter whereas $x$-prediction produces temporally smooth results.

\begin{table}[t]
  \centering
  \caption{Ablation study on flow matching prediction targets on GRAB~\cite{taheri2020grab}. $x$-prediction yields lower angular jerk and hand acceleration while maintaining competitive interaction quality. Acc$_g$/Acc$_l$: global/local acceleration (see Sec.~\ref{sec:supp_metrics}).}
  \label{tab:ablation_prediction_target}
  \resizebox{\linewidth}{!}{%
    \begin{tabular}{@{}l|cc|cccc@{}}
      \toprule
      Prediction Target
      & Jerk (Pos.) [$m/s^3$] ($\downarrow$)
      & Jerk (Ang.) [$rad/s^3$] ($\downarrow$)
      & Acc$_g$ (Pos.) [$m/s^2$] ($\downarrow$)
      & Acc$_l$ (Pos.) [$m/s^2$] ($\downarrow$)
      & Acc$_g$ (Rot.) [$rad/s^2$] ($\downarrow$)
      & Acc$_l$ (Rot.) [$rad/s^2$] ($\downarrow$) \\
      \midrule
      $v$-prediction        & 4.9721 & 137.9271 & 0.2523 & 0.1232 & 0.9567 & 3.2414 \\
      $x$-prediction (Ours) & \textbf{0.0438} & \textbf{0.4071} & \textbf{0.0204} & \textbf{0.0055} & \textbf{0.0835} & \textbf{0.0339} \\
      \bottomrule
    \end{tabular}
  }
\end{table}

\section{Hand-object data reconstruction details}
\label{sec:supp_data_reconstruction}
\begin{figure}[t]
    \centering
    \includegraphics[width=\linewidth]{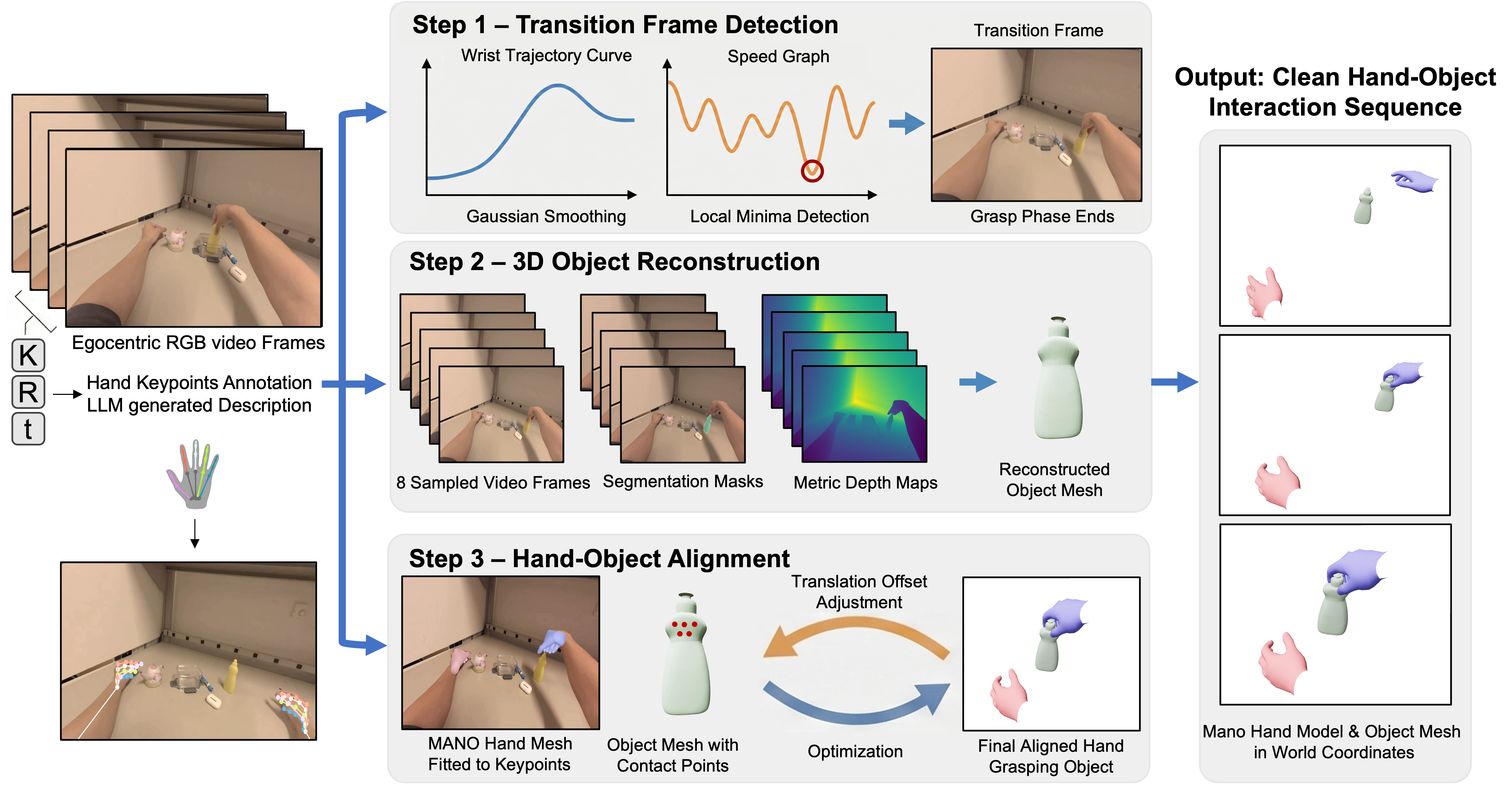}
    \caption{\textbf{Detailed hand-object data reconstruction pipeline.} Given an egocentric RGB video with tracked hand keypoints and camera parameters, we reconstruct HOI data in three steps. (1)~\emph{Transition frame detection}: we smooth wrist trajectories and identify the grasp-to-manipulation transition via local speed minima and direction change. (2)~\emph{3D object reconstruction}: we segment the target object with SAM3~\cite{carion2025sam}, estimate metric depth with DepthAnything3~\cite{lin2025depth}, and reconstruct a mesh with SAM3D~\cite{chen2025sam} using pre-transition frames where the object is static. (3)~\emph{Hand-object alignment}: we fit MANO hand meshes via inverse kinematics and optimize an object translation offset to satisfy fingerpad contact and non-penetration constraints at the transition frame, then propagate the alignment to all frames.}
    \label{fig:data_reconstruction_supp}
\end{figure}

The goal of this pipeline is to reconstruct high-fidelity hand-object interaction data from large-scale egocentric videos~\cite{hoque2025egodex}, which provides large-scale training data for pretraining the grasping model. This section provides the full algorithmic details, loss formulations, and hyperparameters of the reconstruction pipeline summarized in Sec.~\ref{subsec:data_reconstruction} of the main paper. An overview of the pipeline is illustrated in Fig.~\ref{fig:data_reconstruction_supp}.

\subsection{Step 1: transition frame detection}
To separate the grasp phase from the manipulation phase, we detect a transition frame using wrist motion cues.
We first smooth the wrist trajectory with a Gaussian kernel~\cite{chung2020gaussian} ($\sigma{=}2.0$) and compute wrist speed as the displacement between consecutive frames.
Candidate transitions are identified at time steps where the speed reaches a local minimum within a 1-second window and the orientation changes by more than $30^\circ$ in the window.

\subsection{Step 2: 3D object reconstruction}
We use an LLM (GPT-4o~\cite{achiam2023gpt}) to extract the label of the object of interest from the given text prompt.
The object is stable during the grasping phase, allowing us to reconstruct its 3D geometry from the video start frame to the transition frame.
We uniformly sample eight frames from this interval and apply SAM3~\cite{carion2025sam} to segment the target object using the specified label. For each sampled frame, the highest-confidence mask is selected. In cases where SAM3 fails to produce a valid segmentation, we examine nearby frames within the same interval to recover a suitable mask.

Since SAM3D~\cite{chen2025sam} reconstructs geometry from a single RGB image but does not recover metric scale, we first estimate metric depth for the initial frame using DepthAnything3~\cite{lin2025depth} and back-project it into a 3D point map using the known camera parameters. We then apply the segmentation mask to extract the object point map, which provides the metric reference. Given this metrically scaled point map and the initial RGB frame, we reconstruct the object mesh with SAM3D. Given the pose of the RGB image, we further transform the reconstructed metric mesh from SAM3D into the transition frame's world coordinates using the known camera extrinsics.

\subsection{Step 3: hand-object alignment}
Despite sequential video observations from multiple cameras in EgoDex~\cite{hoque2025egodex}, misalignment persists between the reconstructed object and the tracked hand keypoints in the world frame. We align the hand and object based on two hypotheses: (i)~the fingerpads should be in contact (at least three fingers including the thumb) with the object surface at the moment of grasp completion, and (ii)~no deep penetration should occur.

At the transition frame between the grasping and manipulation stages, we first fit the MANO hand mesh from the tracked hand keypoints using optimization-based inverse kinematics (IK)~\cite{kim2025pyroki}, which allows us to extract the fingerpad vertices from the mesh. We then estimate an optimal object translation offset $\Delta\boldsymbol{p}^o\in\mathbb{R}^3$ that minimizes a weighted fingerpad-to-object surface discrepancy at the transition frame by gradient descent. If the resulting grasp does not satisfy the contact hypotheses, we further fine-tune the MANO parameters correction $\Delta\boldsymbol{\theta}$ with:
\begin{equation}
\min_{\Delta\boldsymbol{p}^o, \Delta\boldsymbol{\theta}} \;
\mathcal{L}_{\text{dist}}+\lambda_{\text{pen}}\mathcal{L}_{\text{pen}}+\lambda_{\text{reg}}\mathcal{L}_{\text{reg}},
\end{equation}
where $\Delta\boldsymbol{p}^o\in\mathbb{R}^{3}$ is the object translation offset, $\Delta\boldsymbol{\theta}\in\mathbb{R}^{24}$ is the MANO parameters correction in PCA space, $\mathcal{L}_{\text{dist}}$ penalizes MANO fingerpad to object surface distances, $\mathcal{L}_{\text{pen}}$ penalizes negative signed distances of fingerpad vertices, and $\mathcal{L}_{\text{reg}}$ regularizes deviation from the IK-fitted pose.

Finally, we propagate both the estimated object translation offset $\Delta\boldsymbol{p}^o$ and the optimized MANO pose correction $\Delta\boldsymbol{\theta}$ at the transition frame to all frames prior to the transition, resulting in aligned hand-object data for the entire grasping phase.

\subsection{Reconstruction quality validation on HOT3D}
\label{subsec:supp_recon_validation}

We validate the EgoDex reconstruction pipeline on HOT3D, which provides ground-truth object meshes and 6-DoF poses, by running the same pipeline on HOT3D's egocentric RGB streams and comparing against the released annotations. Table~\ref{tab:supp_recon_quality} summarises the result over the public training split.

\begin{table}[t]
  \caption{Geometric and pose accuracy of our EgoDex reconstruction pipeline measured on HOT3D's ground-truth meshes and 6-DoF poses.}
  \label{tab:supp_recon_quality}
  \centering
  \footnotesize
  \begin{tabular}{@{}l|ccc@{}}
    \toprule
    Metric                  & Mean      & Median    & Std \\
    \midrule
    Chamfer distance        & 9.8\,mm   & 6.5\,mm   & 21.8\,mm \\
    F-score @ 5\,cm         & 96.5\%    & 100.0\%   & 10.3\% \\
    Position error          & 5.1\,cm   & 3.6\,cm   & 6.4\,cm \\
    Mask IoU                & 0.70      & 0.78      & 0.24 \\
    \bottomrule
  \end{tabular}
\end{table}

A median Chamfer distance of 6.5\,mm and a mean F-score@5cm of 96.5\% indicate that nearly all reconstructed surface points lie within 5\,cm of the ground truth, with sub-centimeter median geometric deviation. The median 6-DoF position error of 3.6\,cm is sufficient for initializing the grasping stage, and the upstream SAM3 segmentation (mask IoU median 0.78) confirms reliable object isolation in the egocentric stream.

\subsection[Effect of EgoDex pretraining]{\rev{Effect of EgoDex pretraining}}
\label{sec:supp_egodex}

\rev{We fix the manipulation generator and vary only whether the grasp prefix comes from an EgoDex-pretrained model.
Besides GE, Phy, and SR on GRAB, we report the wrist velocity across the stage boundary (TV), the handedness accuracy (HA), which is the fraction of clips whose first manipulation frame engages the ground-truth hand, and the real-robot grasp success (RG) over 10 separate squeezing trials.
Pretraining lowers GE and TV and raises HA (Table~\ref{tab:supp_egodex}).
Phy, SR, and RG show no clear difference, and all models are trained with a single seed.}

\begin{table}[t]
  \revcolor
  \caption{Effect of EgoDex pretraining. Only the grasp prefix differs. RG is measured on the real robot in a separate set of 10 squeezing trials, and all other metrics on GRAB.}
  \label{tab:supp_egodex}
  \centering
  \footnotesize
  \begin{tabular}{@{}l|cccccc@{}}
    \toprule
    Grasping prior & GE [$m$] ($\downarrow$) & TV [$m/s$] ($\downarrow$) & HA [\%] ($\uparrow$) & Phy [\%] ($\uparrow$) & SR [\%] ($\uparrow$) & RG (of 10) ($\uparrow$) \\
    \midrule
    No EgoDex pretraining    & 0.10          & 0.583          & 96.34          & \textbf{91.06} & 51.38          & 6 \\
    EgoDex-pretrained (ours) & \textbf{0.06} & \textbf{0.340} & \textbf{99.02} & 90.90          & \textbf{55.96} & 6 \\
    \bottomrule
  \end{tabular}
\end{table}

\section{Dataset preprocessing details}
\label{sec:supp_dataset}

This section details our preprocessing and annotation procedures for each dataset used in Sec.~\ref{subsec:datasets} of the main paper.

\subsection{GRAB}

GRAB~\cite{taheri2020grab} is a comprehensive full-body grasping and manipulation dataset containing 1{,}335 sequences of human interactions with 51 everyday objects. Each sequence provides high-quality 3D motion capture data for both the human body and hands, together with accurate object meshes and trajectories. Although GRAB does not include scene information, we include it in our evaluation due to the high fidelity of its hand-object interaction data and the complexity of the manipulation actions it contains.

Following~\cite{christen2024diffh2o}, we focus exclusively on the hand-object interaction components and discard full-body motion information. We adopt the dataset split (1{,}125 train / 210 test) and textual action descriptions introduced in~\cite{christen2024diffh2o}. The raw motion capture data is downsampled from ${\sim}120$\,Hz to 20\,Hz.

\subsection{HOT3D}

For training our semantically grounded and scene-aware manipulation model, there is currently no dataset that simultaneously satisfies all of the following requirements:
(i)~high-quality 3D annotations for hand-object interactions~\cite{chao2021dexycb, taheri2020grab},
(ii)~high-fidelity real-world scene capture comparable to large-scale indoor datasets~\cite{yeshwanth2023scannet++, ramakrishnan2021habitat}, and
(iii)~explicit action-level semantic descriptions.

HOT3D~\cite{banerjee2025hot3d} is close to meeting these requirements. The dataset provides accurate 3D hand and object pose annotations synchronized with egocentric videos, enabling detailed analysis of fine-grained manipulation activities.

In our experiment, we use the publicly available training split of HOT3D, as the ground-truth annotations for the official test set have not been released and are primarily intended for benchmark evaluation.
The training split consists of 136 recordings collected from nine subjects, with each recording lasting approximately two minutes and containing multiple hand-object interactions.

To obtain scene representations, we reconstruct each recording using 3DGS~\cite{kerbl20233d,lv2025photoreal} from the recorded egocentric videos captured by the Project Aria glasses~\cite{engel2023project}. We then automatically extract hand-object interaction clips by detecting object motion and temporally segmenting the sequences accordingly. Specifically, each interaction clip is divided into a grasping stage and a manipulation stage. The grasping stage is defined as the interval from the end of the previous interaction to the moment of grasp completion. The manipulation stage covers the remaining period where the hand actively manipulates the object.

Following this procedure, we obtain 1{,}802 interaction clips in total, split into 1{,}441 for training and 361 for testing, where the test set contains interactions involving unseen objects.
Finally, we use OpenAI GPT-4o~\cite{achiam2023gpt} to generate natural-language descriptions for each extracted interaction clip, including the objects involved and the actions performed.

\noindent\textbf{HOT3D evaluation noise and LatentHOI reproduction.}
Compared to GRAB's mocap supervision, HOT3D's evaluation regime carries three additional sources of noise: roughly 20\% of object poses are missing in the public training split, the initial hand pose is estimated by an off-the-shelf tracker rather than mocap-tracked, and the 3DGS scenes contain reconstruction artifacts (mesh holes, texture seams, transient floaters). These factors compound at the simulator level: the Real Mocap upper bound on HOT3D physics-simulation SR is only 6.67\% (Table~\ref{tab:main_results}), so absolute SR is dominated by simulator and supervision noise rather than by generator quality.

LatentHOI~\cite{li2025latenthoi} collapses to 0\% physics-simulation SR on HOT3D (Table~\ref{tab:main_results}), which we attribute to its autoregressive latent decoder. Small early-frame errors are amplified through the rollout, and under HOT3D's noisier initialization the latent decoder converges on near-static, contact-avoiding sequences that satisfy the heuristic Phy criterion (the object stays above the ground) without any meaningful contact. To rule out a broken retraining as the cause, we also evaluated LatentHOI under its original CLIP text encoder and observed the same HOT3D collapse. On GRAB, LatentHOI's SR (28.44\%) is consistent with the trends reported by the original authors, indicating that the CLIP$\to$T5 swap applied uniformly to all baselines for fairness is not the source of the HOT3D failure.

\section{Physics simulation analysis}
\label{sec:supp_simulation}

This section provides additional details on the physics simulation experiments described in Sec.~\ref{subsec:metrics} of the main paper.

We evaluate the realizable physical feasibility of generated HOI sequences in Isaac Gym~\cite{makoviychuk2021isaac}. The evaluation pipeline consists of three steps: (1)~retargeting the generated MANO~\cite{MANO:SIGGRAPHASIA:2017} hand trajectories to the Allegro Hand~\cite{allegrohand} via inverse kinematics~\cite{kim2025pyroki}, (2)~executing the retargeted motions using a physics-based tracking controller~\cite{liu2025dextrack} that produces robot joint torques, and (3)~evaluating whether the object remains stably grasped and the task is completed. We describe each step in detail below.

\noindent\textbf{Retargeting.}
Given the generated MANO hand trajectories, we first convert the MANO parameters to 21 keypoint positions per hand at each frame via forward kinematics. Since the Allegro Hand is approximately 1.6$\times$ larger than the human hand, we uniformly scale the MANO keypoints accordingly to match the robot hand size. We then solve for the 16-DOF Allegro joint angles and the 6-DOF wrist pose via inverse kinematics using pyroki~\cite{kim2025pyroki}, a JAX-based least-squares IK solver.
(The Allegro Hand has 4 fingers with 4 DOF each = 16 DOF, while MANO models 5 fingers with 3 PCA-coupled DOF each plus a 6-DOF wrist root, hence the structural mismatch we resolve via the offset described below.) The optimization minimizes a weighted combination of cost terms: (i)~a \emph{local alignment} cost that preserves relative inter-joint distances and angles to maintain hand structure, (ii)~a \emph{global alignment} cost that matches the robot link positions to the target keypoints, (iii)~\emph{wrist position and rotation} costs that anchor the robot wrist to the MANO wrist pose using translation error and SO(3) geodesic distance respectively, and (iv)~\emph{temporal smoothness} costs on both the joint angles and the root trajectory to suppress jitter. Joint angle limits from the Allegro URDF are enforced as hard constraints. Since the Allegro Hand has four fingers while MANO models five, we approximate the pinky targets by offsetting the ring finger keypoints laterally.

\noindent\textbf{Tracking.}
We use DexTrack~\cite{liu2025dextrack}, a reinforcement learning-based tracking controller, to execute the retargeted trajectories under physics simulation. The controller is trained with Proximal Policy Optimization (PPO) in Isaac Gym~\cite{makoviychuk2021isaac} across thousands of parallel environments. The policy observes the current hand joint positions and velocities, fingertip states, object pose, and the residual between the current state and the reference trajectory at the current and future timesteps. It outputs cumulative residual position targets for the 16 actuated hand joints, which are tracked by low-level PD controllers. The reward function encourages (i)~minimizing the distance between the fingertips and the object surface, (ii)~tracking the reference hand pose with separate coefficients for global translation, wrist rotation, and finger joint angles, (iii)~matching the target object position and orientation, and (iv)~smooth joint velocity profiles. To improve robustness, we apply domain randomization over hand joint stiffness and damping, hand and object masses, friction coefficients, and observation and action noise during training.

\begin{figure}[t]
    \centering
    \includegraphics[width=\linewidth]{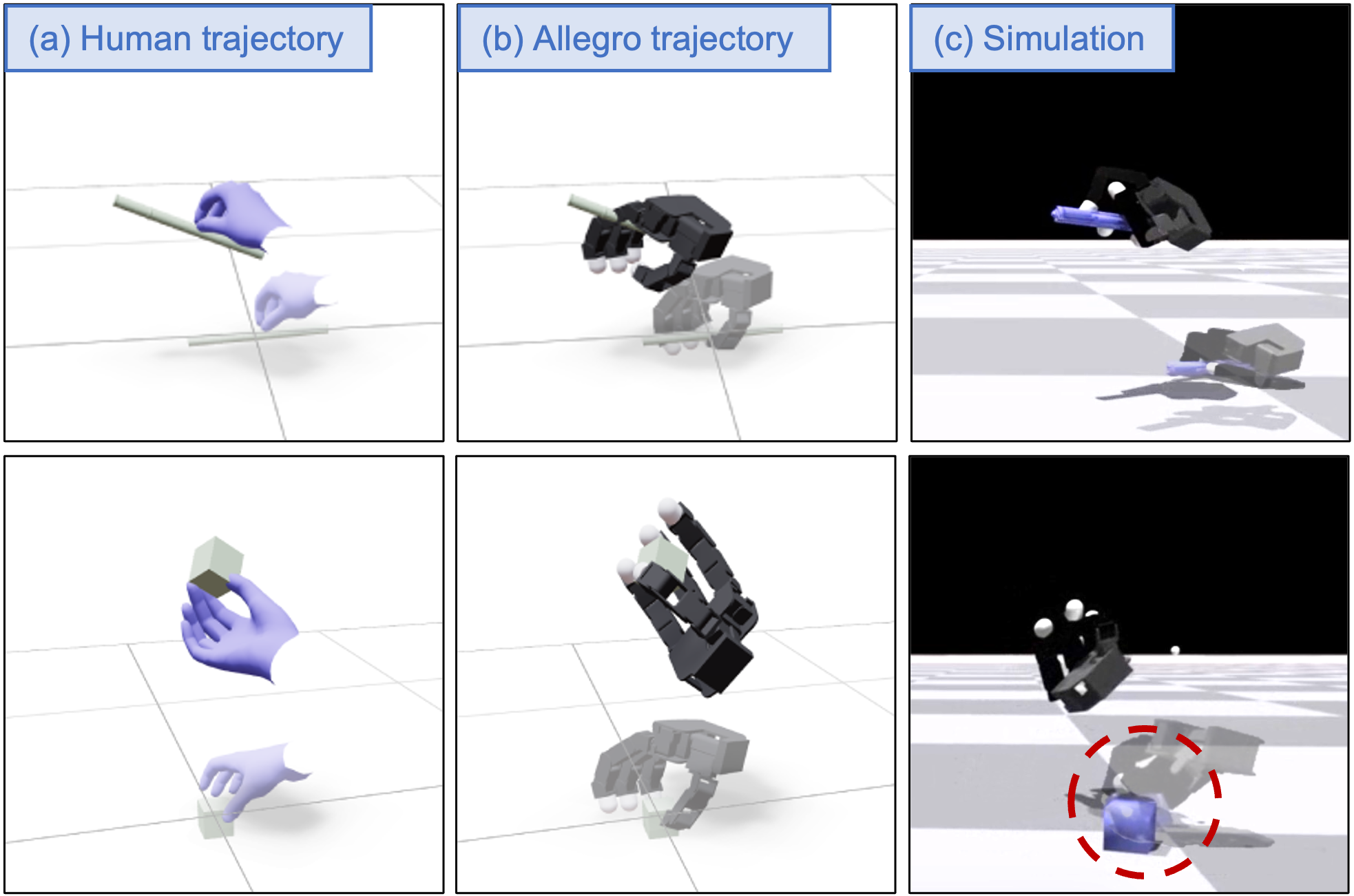}
    \caption{Comparison of HOI trajectories: (a)~kinematic human hand trajectory, (b)~retargeted Allegro Hand trajectory, and (c)~physics simulation in Isaac Gym~\cite{makoviychuk2021isaac}. Although the kinematic trajectories in (a) and (b) appear plausible, executing them under physics simulation reveals failures: the object detaches and slips from the grasp (bottom row, red circle), indicating that kinematically feasible motions do not guarantee physically stable interactions.}
    \label{fig:simulation_supp}
\end{figure}

\noindent\textbf{Evaluation.}
As shown in Fig.~\ref{fig:simulation_supp}, although the kinematic trajectories of both the human hand (a) and the retargeted Allegro Hand (b) appear visually plausible, executing them under physics simulation (c) can reveal failures: the object may detach or slip from the grasp due to insufficient contact forces, incorrect friction modeling, or dynamically unstable grasp configurations. This demonstrates that kinematic feasibility alone does not guarantee physically stable interactions, motivating the use of physics simulation as a more rigorous evaluation protocol.

For each method, we run 10 simulation trials per test sequence and report the average success rate and holding time. An episode is considered successful if the object remains within 5\,cm of its target position while maintaining finger contact throughout the trajectory. All simulation parameters follow the default settings from DexTrack~\cite{liu2025dextrack}.

\section{Real-world experiment details}
\label{sec:supp_realworld}

This section provides additional details on the real-world dexterous manipulation experiments showcased in the main paper.

\noindent\textbf{Hardware Setup.}
Our platform consists of two Franka Emika Panda robotic arms~\cite{haddadin2024franka}, each equipped with an Allegro Hand v5~\cite{allegrohand}.
The 3D scene is reconstructed offline prior to each experiment: we capture a multi-view video sequence using a MetaCam~\cite{metacam} scanner and reconstruct the scene via Gaussian-LIC~\cite{lang2025gaussian}, yielding a 3DGS representation consistent with our training pipeline.

\noindent\textbf{Retargeting Pipeline.}
Given the generated MANO hand trajectories, we decompose the retargeting into two stages:
(1)~\emph{Arm retargeting}: the generated MANO wrist 6D pose at each frame is used as the end-effector target for the Franka arm. We solve the 7-DOF joint angles using the Franka built-in inverse kinematics solver.
(2)~\emph{Hand retargeting}: the MANO finger articulations are retargeted to the 16-DOF Allegro Hand joint angles using pyroki~\cite{kim2025pyroki}, following the same procedure described in Sec.~\ref{sec:supp_simulation}.
The retargeted trajectories are further refined offline by the DexTrack~\cite{liu2025dextrack} sim-to-real policy, which was trained entirely in Isaac Gym~\cite{makoviychuk2021isaac} and deployed without additional fine-tuning on the real robot.

\noindent\textbf{Execution.}
The refined joint trajectories are executed in an open-loop fashion via a standard joint impedance controller on both the Franka arms and the Allegro Hands at 1\,kHz.
Since the execution is open-loop, no real-time object state feedback is required during task execution.

\noindent\textbf{Evaluation.}
We evaluate the system on four contact-rich household manipulation tasks: drinking from a cup, pouring liquid between containers of different sizes, tilting a container, and squeezing dressing\rev{, with 10 trials per task (Table~\ref{tab:realworld})}. We refer the reader to Fig.~\ref{fig:real_world} in the main paper and the supplementary video for visual results.

\noindent\textbf{Task diversity (joint-angle analysis).}
The four real-world tasks are not kinematically identical. Pouring and tilting demand distinct wrist rotation ranges and grasp-stability constraints. Drinking is a compound lift-tilt-return trajectory. Squeezing requires coordinated finger flexion with distinct joint targets and force profiles.

\rev{\noindent\textbf{Phase decomposition of the tasks.}
Each task uses a single grasp-to-manipulation transition.
In squeezing, the finger closure that establishes contact belongs to the grasping stage, and the compression belongs to the manipulation stage.
Multi-phase tasks such as opening a bottle lid and deformable objects such as a toothpaste tube are outside the scope of this work.}

\section{Scene representation details and analysis}
\label{sec:supp_scene}

This section provides additional details on the scene representation described in Sec.~\ref{subsec:two_stage_generation} of the main paper.

\begin{figure}[t]
    \centering
    \includegraphics[width=0.9\linewidth]{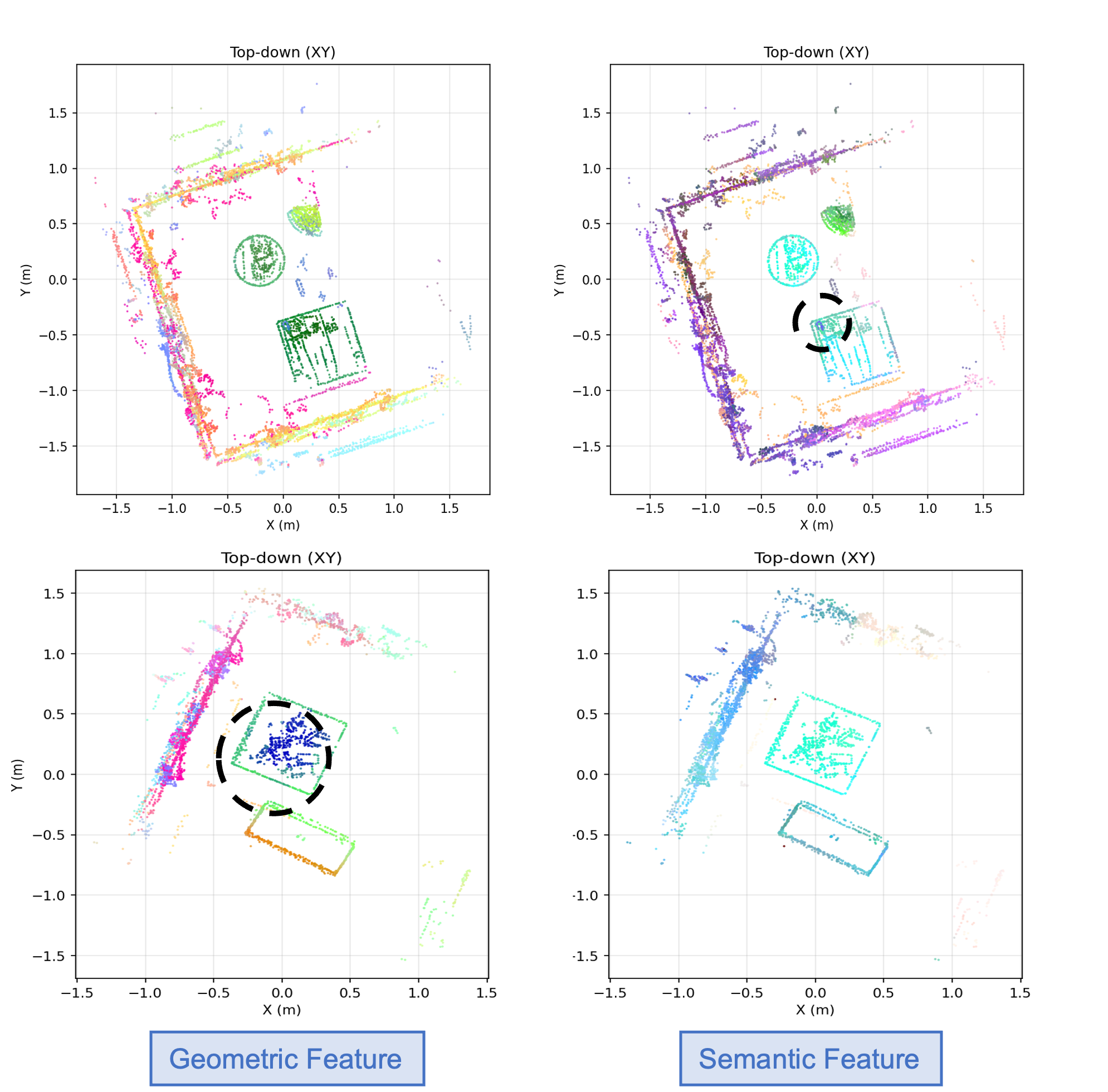}
    \caption{\textbf{Scene feature analysis.} Top-down views of two scenes comparing geometric features (Concerto~\cite{zhang2025concerto}, left) and semantic features (SceneSplat~\cite{li2025scenesplat}, right), colored by PCA of the respective feature spaces. Black dashed circles highlight where geometric features successfully distinguish objects from the supporting surface (bottom-left) and where semantic features successfully discriminate semantically distinct regions (top-right). This complementarity motivates our bidirectional fusion.}
    \label{fig:scene_analysis}
\end{figure}

Fig.~\ref{fig:scene_analysis} visualizes the geometric features (Concerto~\cite{zhang2025concerto}) and semantic features (SceneSplat~\cite{li2025scenesplat}) of two example scenes in top-down view, colored by PCA of the respective feature spaces.
The geometric features (left column) primarily encode spatial properties such as surface normals, curvature, and local point distributions. As highlighted by the black dashed circles, geometric features can distinguish objects on a table from the table surface itself based on their different spatial structures.
The semantic features (right column) encode language-aligned semantics, grouping points by object category or functional role. Points on the same object share coherent representations regardless of spatial location, while semantically different objects are clearly separated (black dashed circles).
This complementarity motivates our bidirectional fusion: geometric features provide precise spatial grounding for localization and collision avoidance, while semantic features enable the model to identify task-relevant objects for text-conditioned generation.

\subsection[Scene-conditioning granularity]{\rev{Scene-conditioning granularity}}
\label{sec:supp_scene_granularity}

\rev{On HOT3D, we compare no scene conditioning, global-only, local-only, and global-plus-local conditioning, with everything else fixed.
In addition to GE and HA (Sec.~\ref{sec:supp_egodex}), we report the hand-scene penetration depth (PD, Sec.~\ref{sec:supp_clutter}).
The global token mainly improves interaction placement and hand selection (GE, HA), and the local tokens mainly improve hand-object contact (IV, CR).
Combining both gives the best GE, HA, Phy, and PD, although PD differs only slightly across variants (Table~\ref{tab:supp_scene_granularity}).}

\begin{table}[t]
  \revcolor
  \caption{Scene-conditioning granularity on HOT3D. GE: grasp error at the transition frame. HA: handedness accuracy. PD: hand-scene penetration depth.}
  \label{tab:supp_scene_granularity}
  \centering
  \footnotesize
  \begin{tabular}{@{}l|cccccc@{}}
    \toprule
    Scene conditioning & IV [$cm^3$] ($\downarrow$) & CR [\%] ($\uparrow$) & Phy [\%] ($\uparrow$) & GE [$m$] ($\downarrow$) & HA ($\uparrow$) & PD [$mm$] ($\downarrow$) \\
    \midrule
    None                  & 3.26          & 2.27          & 16.47          & 0.20          & 0.83          & 2.08 \\
    Global only           & 3.30          & 2.09          & 16.63          & 0.14          & 0.88          & 1.97 \\
    Local only            & \textbf{3.22} & \textbf{2.33} & 16.10          & 0.19          & 0.85          & 1.98 \\
    Global + local (ours) & 3.36          & 2.14          & \textbf{17.15} & \textbf{0.11} & \textbf{0.92} & \textbf{1.95} \\
    \bottomrule
  \end{tabular}
\end{table}

\subsection[Hand-scene penetration under scene clutter]{\rev{Hand-scene penetration under scene clutter}}
\label{sec:supp_clutter}

\rev{We split the HOT3D test sequences into three equal groups by a clutter score, the number of occupied scene voxels (8.3\,cm) within 0.5\,m of the ground-truth object trajectory, excluding the target object.
Penetration is measured from MANO vertices to the MPS semidense point cloud released with HOT3D, which contains only the static scene.
We report the mean penetration depth over penetrating vertex-frames and a GT-tolerance success rate, the fraction of clips whose maximum per-frame penetration stays below 6.66\,mm (the 95th percentile of the ground truth).
The ground truth is evaluated with the same pipeline as a reference.
FlowHOI stays within 0.03\,mm of the ground-truth penetration depth in every group, and its success-rate gap to the ground truth shrinks from 6.7 to 0.9 points as clutter increases (Table~\ref{tab:supp_clutter}).}

\begin{table}[t]
  \revcolor
  \caption{Hand-scene penetration against the HOT3D MPS static point cloud, stratified by scene clutter.}
  \label{tab:supp_clutter}
  \centering
  \footnotesize
  \begin{tabular}{@{}ll|ccc@{}}
    \toprule
    Metric & Source & Low ($n{=}120$) & Medium ($n{=}120$) & High ($n{=}120$) \\
    \midrule
    \multirow{2}{*}{Penetration depth [$mm$] ($\downarrow$)}  & FlowHOI      & 1.97 & 1.94 & 2.04 \\
                                                             & Ground truth & 1.94 & 1.94 & 2.02 \\
    \midrule
    \multirow{2}{*}{GT-tolerance success rate [\%] ($\uparrow$)} & FlowHOI      & 70.0 & 80.8 & 85.8 \\
                                                                & Ground truth & 76.7 & 85.0 & 86.7 \\
    \bottomrule
  \end{tabular}
\end{table}

\section{Evaluation metric details}
\label{sec:supp_metrics}

This section provides detailed definitions and implementation details of the evaluation metrics reported in Sec.~\ref{subsec:metrics} of the main paper.

\subsection{Physical interaction quality}

Let $\mathcal{M}^t_h$ and $\mathcal{M}^t_o$ denote the hand and object meshes at frame $t$, respectively.
For a hand mesh vertex $\mathbf{v}$, we denote by $d(\mathbf{v}, \mathcal{M}^t_o)$ the signed distance to the object surface, where negative values indicate penetration.
All physical interaction metrics (IV, ID, CR, IVU) are computed independently for each hand and then averaged across hands.

\noindent\textbf{Interpenetration Volume (IV).}
Interpenetration volume measures the volumetric intersection between the hand and object meshes:
\begin{equation}
\mathrm{IV}_t = \mathrm{Vol}\!\left(\mathcal{M}^t_h \cap \mathcal{M}^t_o\right),
\end{equation}
and is averaged over time:
\begin{equation}
\mathrm{IV} = \frac{1}{T} \sum_{t=1}^{T} \mathrm{IV}_t.
\end{equation}

\noindent\textbf{Interpenetration Depth (ID).}
Interpenetration depth is defined as the mean penetration distance across all penetrating hand vertices:
\begin{equation}
\mathrm{ID}_t =
\frac{1}{|\mathcal{V}^t_{\mathrm{pen}}|}
\sum_{\mathbf{v} \in \mathcal{V}^t_{\mathrm{pen}}}
\max\!\left(0, -d(\mathbf{v}, \mathcal{M}^t_o)\right),
\end{equation}
where $\mathcal{V}^t_{\mathrm{pen}} = \{\mathbf{v} \in \mathcal{V}_h \mid d(\mathbf{v}, \mathcal{M}^t_o) < 0\}$ is the set of penetrating vertices at frame $t$.
ID is temporally averaged over all frames.

\noindent\textbf{Contact Ratio (CR).}
The contact ratio quantifies the proportion of hand vertices that are close to the object surface:
\begin{equation}
\mathrm{CR}_t =
\frac{1}{|\mathcal{V}_h|}
\sum_{\mathbf{v} \in \mathcal{V}_h}
\mathbb{I}\!\left(d(\mathbf{v}, \mathcal{M}^t_o) \le \delta\right),
\end{equation}
where $\delta$ is a fixed distance threshold (set to $5$\,mm in all experiments).
The final CR score is obtained by averaging over interaction frames (frames where hand-object contact occurs).

\noindent\textbf{Interpenetration Volume per Contact Unit (IVU).}
Since low penetration can be trivially achieved by avoiding contact, we additionally report interpenetration volume normalized by the contact area at each frame.
For each contact frame $t$ (where $\mathrm{CR}_t > 0$), we compute the per-frame IVU as:
\begin{equation}
\mathrm{IVU}_t
=
\frac{\mathrm{IV}_t}{A^t_{\mathrm{contact}}},
\quad
A^t_{\mathrm{contact}} = \mathrm{CR}_t \cdot \mathrm{Area}(\mathcal{M}_h),
\end{equation}
and the final IVU is the mean over all contact frames:
\begin{equation}
\mathrm{IVU}
=
\frac{1}{|\mathcal{T}_c|}
\sum_{t \in \mathcal{T}_c}
\mathrm{IVU}_t,
\end{equation}
where $\mathcal{T}_c = \{t \mid \mathrm{CR}_t > 0\}$ is the set of frames with non-zero contact.

\subsection{Motion quality}

\noindent\textbf{Action Recognition Accuracy (AR).}
Following IMoS~\cite{ghosh2023imos}, we pretrain a 4-layer Transformer encoder with a \texttt{[CLS]} classification head on each dataset's training split (29 action classes on GRAB, 14 on HOT3D), holding it fixed thereafter and using it to score every method's generated trajectory under identical conditions. The classifier consumes the bimanual MANO joint-position sequence, augmented with the object-pose sequence on HOT3D where scene context is available. AR is the top-1 classification accuracy on the test split:
\begin{equation}
\mathrm{AR} = \frac{1}{N}\sum_{i=1}^{N}\mathbb{I}\!\left(\arg\max f_{\mathrm{act}}(\hat{\mathbf{x}}_i) = y_i\right).
\end{equation}

\noindent\textbf{Sample Diversity (SD).}
For each conditioning input, we generate $K$ motion samples and compute the mean pairwise $\ell_2$ distance between their hand joint trajectories, normalized by the number of frames:
\begin{equation}
\mathrm{SD}
=
\frac{2}{K(K-1)}
\sum_{1 \le i < j \le K}
\frac{1}{T'}
\left\|
\mathbf{h}^{(i)} - \mathbf{h}^{(j)}
\right\|_2,
\end{equation}
where $\mathbf{h}^{(i)} \in \mathbb{R}^{2J \cdot 3 \cdot T'}$ is the flattened joint position trajectory of the $i$-th sample. We use $K=4$ in our experiments.

\noindent\textbf{Overall Diversity (OD).}
Overall diversity is computed as the mean pairwise $\ell_2$ distance between hand joint trajectories across all generated test samples, normalized by the number of frames:
\begin{equation}
\mathrm{OD}
=
\frac{2}{S(S-1)}
\sum_{1 \le i < j \le S}
\frac{1}{T'}
\left\|
\mathbf{h}_i - \mathbf{h}_j
\right\|_2,
\end{equation}
where $S$ is the total number of generated motions.

\subsection{Realizable physical feasibility}

\noindent\textbf{Physical Plausibility (Phy).}
Following LatentHOI~\cite{li2025latenthoi}, physical plausibility is a heuristic, per-sequence binary score.
A sequence is deemed physically plausible if
(i)~at least one hand joint remains within a signed distance of $5$\,mm to the object surface for every frame during the interaction phase, and
(ii)~the object stays above the ground plane throughout the sequence.
The final Phy score is the percentage of test sequences satisfying both criteria.
Note that this metric only checks kinematic proximity and does not account for dynamic stability. As discussed in Sec.~\ref{sec:supp_simulation}, a high Phy score does not necessarily imply physically realizable interactions.

\noindent\textbf{Success Rate (SR).}
For simulation evaluation, a trial is considered successful if the object is lifted above the ground plane by at least $5$\,cm while maintaining contact with the hand. The success rate is the fraction of such sequences:
\begin{equation}
\mathrm{SR} =
\frac{1}{N}
\sum_{n=1}^{N}
s_n,
\end{equation}
where $s_n \in \{0,1\}$ indicates whether the object in the $n$-th trial is lifted off the ground ($>5$\,cm) with sustained hand-object contact.

\noindent\textbf{Holding Time (HT).}
Holding time measures the duration during which objects are stably held in the hand and is averaged over successful trials:
\begin{equation}
\mathrm{HT}
=
\frac{\sum_{n=1}^{N} s_n \, \tau_n}
{\sum_{n=1}^{N} s_n},
\end{equation}
where $\tau_n$ denotes the duration during which the object is stably held in the $n$-th trial.

\subsection{Ablation-specific metrics}

\noindent\textbf{Grasp Error (GE).}
Grasp error measures the mean per-joint Euclidean distance between the generated and ground-truth hand poses at the transition frame $t_g$:
\begin{equation}
\mathrm{GE} = \frac{1}{N_{\text{test}}} \sum_{i=1}^{N_{\text{test}}} \frac{1}{|\mathcal{J}^{(i)}|} \sum_{j \in \mathcal{J}^{(i)}} \left\| \hat{\mathbf{j}}^{(i)}_{t_g,j} - \mathbf{j}^{(i)}_{t_g,j} \right\|_2,
\end{equation}
where $\hat{\mathbf{j}}^{(i)}_{t_g,j}, \mathbf{j}^{(i)}_{t_g,j} \in \mathbb{R}^3$ are the generated and ground-truth 3D positions of the $j$-th MANO joint at the transition frame, and $\mathcal{J}^{(i)}$ is the set of joints belonging to the grasping hand(s), determined by proximity to the object at the ground-truth transition frame.

\noindent\textbf{Final Displacement Error (FDE).}
Final displacement error measures the Euclidean distance between the generated and ground-truth object poses at the end of the manipulation stage:
\begin{equation}
\mathrm{FDE} = \frac{1}{N_{\text{test}}} \sum_{i=1}^{N_{\text{test}}} \left\| \hat{\mathbf{O}}^{(i)}_{N-1} - \mathbf{O}^{(i)}_{N-1} \right\|_2,
\end{equation}
where $\hat{\mathbf{O}}^{(i)}_{N-1}$ and $\mathbf{O}^{(i)}_{N-1}$ denote the generated and ground-truth object poses at the final frame.

\noindent\textbf{Jerk.}
Jerk measures the smoothness of generated object motion as the third derivative of position with respect to time, i.e., the rate of change of acceleration.
Human motion naturally follows a \emph{minimum jerk} principle~\cite{flash1985coordination}, and lower jerk values indicate smoother, more natural trajectories.
Given the object trajectory $\mathbf{p}^t \in \mathbb{R}^3$ over $T$ uniformly sampled frames, we compute the positional jerk via finite differences:
\begin{equation}
\mathrm{Jerk}_{\text{pos}} = \frac{1}{T} \sum_{t=1}^{T} \left\| \frac{d^3 \mathbf{p}^t}{dt^3} \right\|_2,
\end{equation}
where derivatives are approximated using central differences (\texttt{np.gradient}).
We analogously define the angular jerk $\mathrm{Jerk}_{\text{ang}}$ over the object rotation trajectory $\boldsymbol{\theta}^t \in \mathbb{R}^3$ (axis-angle representation).

\noindent\textbf{Hand Acceleration.}
To assess the stability of generated hand motion, we compute acceleration metrics that separately evaluate global wrist movement and local finger articulation.
Given the MANO joint positions $\mathbf{J}^t \in \mathbb{R}^{J \times 3}$ at frame $t$, we decompose them into wrist positions $\mathbf{p}^t_w \in \mathbb{R}^{3}$ and finger positions $\mathbf{p}^t_f \in \mathbb{R}^{(J-1) \times 3}$.
Local finger positions are computed relative to the wrist: $\tilde{\mathbf{p}}^t_f = \mathbf{p}^t_f - \mathbf{p}^t_w$, isolating finger articulation from global wrist movement.
We then compute acceleration via second-order finite differences and report the mean absolute acceleration:
\begin{equation}
\mathrm{Acc}_g^{\text{pos}} = \frac{1}{T}\sum_{t}\left|\frac{d^2 \mathbf{p}^t_w}{dt^2}\right|, \quad
\mathrm{Acc}_l^{\text{pos}} = \frac{1}{T}\sum_{t}\left|\frac{d^2 \tilde{\mathbf{p}}^t_f}{dt^2}\right|,
\end{equation}
where $|\cdot|$ denotes element-wise absolute value and the result is averaged over all spatial components and both hands.
The rotational counterparts $\mathrm{Acc}_g^{\text{rot}}$ and $\mathrm{Acc}_l^{\text{rot}}$ are defined analogously over the wrist and finger axis-angle rotations, respectively.
Separating global and local components ensures that wrist trajectory smoothness and finger articulation stability are evaluated independently.

%
\section{Stage-transition inpainting ablation}
\label{sec:supp_inpainting}

We ablate the soft inpainting mechanism (Sec.~\ref{subsec:two_stage_generation}, the $\mathbf{M}\!\odot\!\mathbf{x}^m_\tau + (1-\mathbf{M})\!\odot\!\mathbf{x}^g_{[0,t_g]}$ replacement during inference) on GRAB. Without inpainting, the manipulation network receives no grasping context, causing velocity discontinuities at the transition frame. Table~\ref{tab:supp_inpainting} reports motion-smoothness metrics: positional jerk ($\mathrm{Jerk}_{\mathrm{pos}}$), angular jerk ($\mathrm{Jerk}_{\mathrm{ang}}$), global wrist acceleration ($\mathrm{Acc}_g^{\mathrm{pos}}$), and grasp error (GE).

\begin{table}[t]
  \caption{Effect of stage-transition soft inpainting on GRAB. Inpainting suppresses high-frequency artifacts at the grasp/manipulation boundary by an order of magnitude.}
  \label{tab:supp_inpainting}
  \centering
  \footnotesize
  \begin{tabular}{@{}l|cccc@{}}
    \toprule
    Setting           & $\mathrm{Jerk}_{\mathrm{pos}}$ ($\downarrow$) & $\mathrm{Jerk}_{\mathrm{ang}}$ ($\downarrow$) & $\mathrm{Acc}_g^{\mathrm{pos}}$ ($\downarrow$) & GE [$m$] ($\downarrow$) \\
    \midrule
    w/o inpainting    & $0.341$           & $3.218$           & $0.157$           & $0.14$ \\
    w/ inpainting     & $\mathbf{0.044}$  & $\mathbf{0.407}$  & $\mathbf{0.020}$  & $\mathbf{0.06}$ \\
    \bottomrule
  \end{tabular}
\end{table}


\section[Single-stage versus two-stage generation]{\rev{Single-stage versus two-stage generation}}
\label{sec:supp_single_stage}

\rev{The single-stage variant generates the same 200-frame sequence in one pass and matches the two-stage model in training data, objective, inference steps, parameter budget, and EgoDex pretraining.
The two-stage model improves the contact ratio (26\% higher on HOT3D), Phy, and interpenetration volume on both datasets, while AR and SR are comparable and the single-stage model is 40\,ms faster (Table~\ref{tab:supp_single_stage}).}

\begin{table}[t]
  \revcolor
  \caption{Single-stage versus two-stage generation. Both variants share the representation, training data, objective, inference steps, parameter budget, and EgoDex pretraining.}
  \label{tab:supp_single_stage}
  \centering
  \footnotesize
  \begin{tabular}{@{}ll|cccccc@{}}
    \toprule
    Dataset & Model & IV [$cm^3$] ($\downarrow$) & CR [\%] ($\uparrow$) & Phy [\%] ($\uparrow$) & AR ($\uparrow$) & SR [\%] ($\uparrow$) & Time [s] ($\downarrow$) \\
    \midrule
    \multirow{2}{*}{GRAB}  & Single-stage     & 11.94          & 6.75          & 89.83          & 0.93          & 54.12          & \textbf{0.12} \\
                           & Two-stage (ours) & \textbf{10.93} & \textbf{6.85} & \textbf{90.90} & \textbf{0.95} & \textbf{55.96} & 0.16 \\
    \midrule
    \multirow{2}{*}{HOT3D} & Single-stage     & 3.53           & 1.70          & 16.42          & \textbf{0.80} & \textbf{6.67}  & \textbf{0.12} \\
                           & Two-stage (ours) & \textbf{3.36}  & \textbf{2.14} & \textbf{17.15} & 0.78          & 5.33           & 0.16 \\
    \bottomrule
  \end{tabular}
\end{table}

\section[Post-grasp dexterity analysis]{\rev{Post-grasp dexterity analysis}}
\label{sec:supp_dexterity}

\rev{We evaluate the 174 of 410 GRAB hand instances with an asymmetric object and more than $30^\circ$ of ground-truth hand-relative rotation.
RRE is the geodesic error of the hand-relative object orientation over post-grasp contact frames, and CT-F1 scores joint-level contact make and release events.
FlowHOI reduces RRE by 44\% while producing 1.11$\times$ the ground-truth rotation range, and it over-segments contact less than DiffH2O (2.5$\times$ vs.\ 4.5$\times$ the ground-truth number of contact events, Table~\ref{tab:supp_dexterity}).}

\begin{table}[t]
  \revcolor
  \caption{Post-grasp dexterity on the 174-instance GRAB reorientation subset.}
  \label{tab:supp_dexterity}
  \centering
  \footnotesize
  \begin{tabular}{@{}l|cc@{}}
    \toprule
    Method & RRE [deg] ($\downarrow$) & CT-F1 ($\uparrow$) \\
    \midrule
    DiffH2O~\cite{christen2024diffh2o} & 83.0          & 0.051 \\
    FlowHOI (ours)                     & \textbf{46.4} & \textbf{0.079} \\
    \bottomrule
  \end{tabular}
\end{table}

\section{Additional qualitative results}
\label{sec:supp_qualitative}

\begin{figure}[t]
    \centering
    \includegraphics[width=\linewidth]{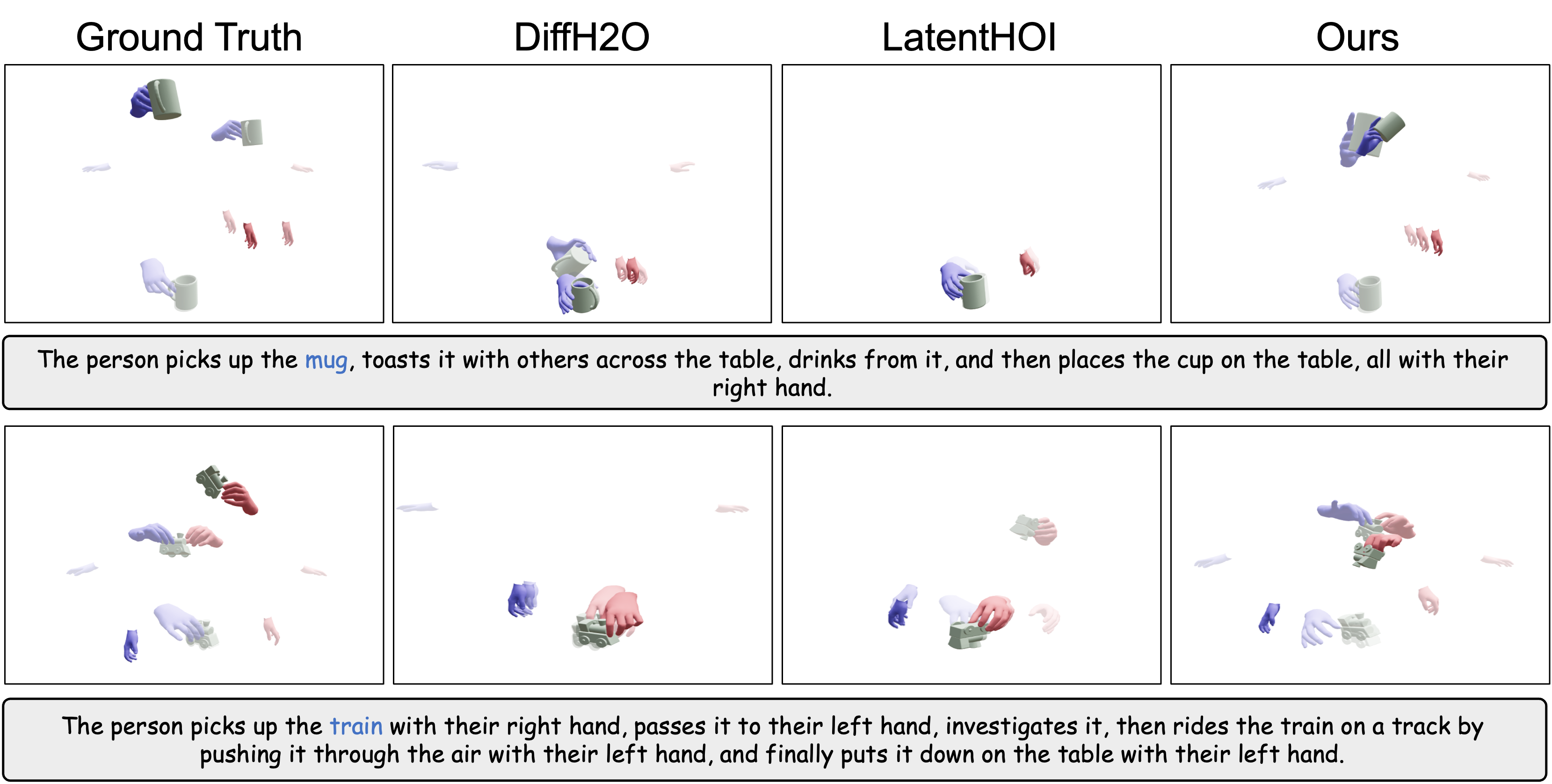}
    \caption{\textbf{Qualitative comparison on GRAB.} Comparison with DiffH2O~\cite{christen2024diffh2o} and LatentHOI~\cite{li2025latenthoi} on diverse actions. Our method produces more physically plausible hand-object interactions with accurate grasping poses and smooth bimanual coordination.}
    \label{fig:qualitative_grab_supp}
\end{figure}

\begin{figure}[t]
    \centering
    \includegraphics[width=\linewidth]{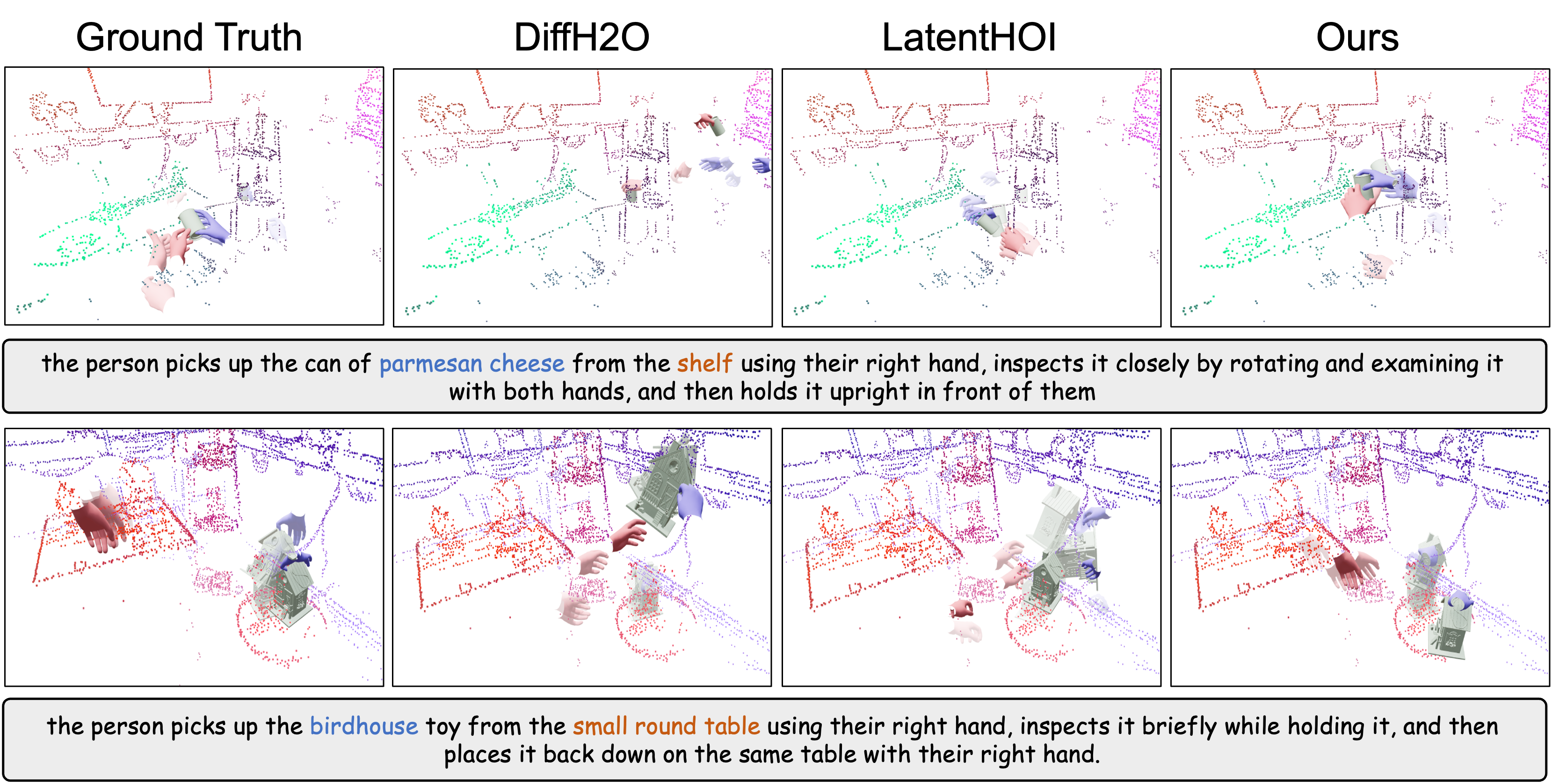}
    \caption{\textbf{Additional qualitative results on HOT3D with scene context.} The generated hand trajectories adapt to the surrounding scene layout, correctly interacting with objects on shelves and tables.}
    \label{fig:qualitative_hot3d_supp}
\end{figure}

We provide additional qualitative results to complement the main paper. Fig.~\ref{fig:qualitative_grab_supp} provides qualitative comparisons with DiffH2O~\cite{christen2024diffh2o} and LatentHOI~\cite{li2025latenthoi} on GRAB~\cite{taheri2020grab}. Fig.~\ref{fig:qualitative_hot3d_supp} presents scene-conditioned results on HOT3D~\cite{banerjee2025hot3d}, demonstrating that the generated trajectories adapt to the surrounding scene layout. \textbf{We strongly encourage the reader to view the supplementary video for animated results.}

\section{Failure cases}
\label{sec:supp_failure}

\begin{figure}[htp]
    \centering
    \includegraphics[width=\linewidth]{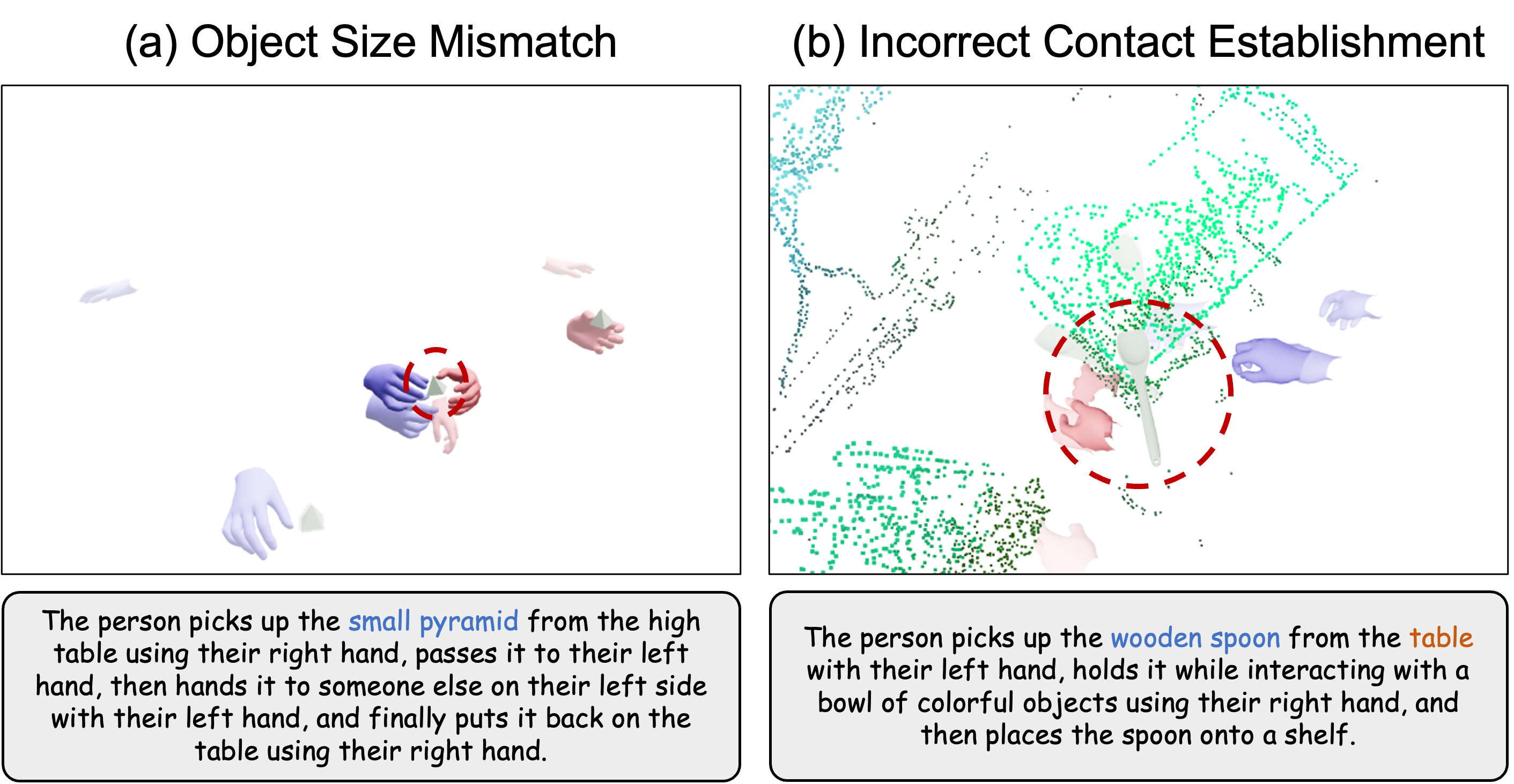}
    \caption{\textbf{Representative failure cases.} Red dashed circles highlight the erroneous regions. (a)~\emph{Object size mismatch}: the small pyramid floats away from the hand due to a mismatch between object size and generated grasp aperture. (b)~\emph{Incorrect contact establishment}: the hand fails to properly wrap around the wooden spoon, resulting in physically implausible finger-object contact. Left: GRAB~\cite{taheri2020grab}, right: HOT3D~\cite{banerjee2025hot3d} with scene context.}
    \label{fig:failure_cases}
\end{figure}

We identify two representative failure modes of our framework (see Fig.~\ref{fig:failure_cases}):

\noindent\textbf{(a) Object Size Mismatch.}
When the object is too large or too small relative to the hand, the generated grasp may not fully enclose the object, resulting in the object appearing to float in the air without stable contact. A potential improvement is to incorporate explicit object size conditioning or adaptive grasp aperture prediction based on object geometry.

\noindent\textbf{(b) Incorrect Contact Establishment.}
The hand and object may fail to establish proper contact during the grasping phase, leading to physically implausible configurations where fingers do not properly wrap around the object surface. Future work could incorporate contact-aware losses or physics-based refinement during inference to ensure proper finger-object contact. Additionally, improving the quality of training data with more accurate contact annotations could also help mitigate this issue.

\end{document}